\documentclass[11pt,fleqn,twoside,titlepage]{newcslarticle}
\newcommand{\ifelse}[2]{#2}
 
\cslreportnumber{CSL Technical Report SRI-CSL-2024-02R3}

\ifelse{\newcommand{\CL}{C{\smaller LARISSA}}}{}
\ifelse{}{}

\acknowledge{SRI Project 101425 in support of
DARPA ANSR Program.\\
Distribution Statement “A” (Approved for Public Release, Distribution Unlimited).}
\usepackage{fancyhdr}
\PassOptionsToPackage{hyphens}{url}
\usepackage[bookmarks=true,hyperfigures=true,colorlinks=true,linkcolor=blue,citecolor=blue,backref=page,pagebackref=false,pdfpagemode=fullscreen,plainpages=false,pdfpagelabels]{hyperref}
\usepackage[all]{hypcap}
\newcommand{\memotwo}[1]{}

\newcommand{\pfd}{\mathit{pfd}}

\usepackage[normalem]{ulem} 
\usepackage{caption,cite,url,relative,latexsym,alltt,nobibhead}
\usepackage[utf8]{inputenc}
\usepackage{array}
\DeclareUnicodeCharacter{25A1}{\ensuremath\Box}
\DeclareUnicodeCharacter{25C7}{\ensuremath\Diamond}
\DeclareSymbolFont{symbolsC}{U}{txsyc}{m}{n}
\DeclareMathSymbol{\strictif}{\mathrel}{symbolsC}{74}

\def\BigLaTeX{{\rm L\kern-.36em\raise.3ex\hbox{\smaller\smaller A}\kern-.15em
    T\kern-.1667em\lower.7ex\hbox{E}\kern-.125emX}}
\def\BoldLaTeX{{\bf L\kern-.36em\raise.3ex\hbox{\smaller\smaller\bf A}\kern-.15em
    T\kern-.1667em\lower.7ex\hbox{E}\kern-.125emX}}
\def\BibTeX{{\rm B\kern-.05em{\sc i\kern-.025em b}\kern-.08em
    T\kern-.1667em\lower.7ex\hbox{E}\kern-.125emX}}

\newlength{\hsbw}
\def\extrawidth{0.5in}

\newlength{\vsbw}
\newcounter{sessioncount}
\newenvironment{session*}{\begin{flushleft}
 \refstepcounter{sessioncount}
 \setlength{\hsbw}{\linewidth}
 \addtolength{\hsbw}{-\arrayrulewidth}
 \addtolength{\hsbw}{-\tabcolsep}
 \begin{tabular}{@{}|c@{}|@{}}\hline 
 \begin{minipage}[b]{\hsbw}
 \vspace*{-.5pt}
 \begin{flushright}
 \rule{0.01in}{.15in}\rule{0.3in}{0.01in}\hspace{-0.35in}
 \raisebox{0.04in}{\makebox[0.3in][c]{\footnotesize \thesessioncount}}
 \end{flushright}
 \vspace*{-.57in}
 \begingroup\small\vspace*{1.0ex}\begin{alltt}}{\end{alltt}\endgroup\end{minipage}\\ \hline 
 \end{tabular}
 \end{flushleft}}
\def\sessionsize{\small}
\def\smallsessionsize{\small}

\newcommand{\exmemo}[1]{}

\newcommand{\comment}[1]{}

\newcommand{\exfootnote}[1]{}
\newlength{\sblen}
\newlength{\overhang}

\def\SetFigFont#1#2#3{\rm}

\newcommand{\excite}[1]{}

\newcommand{\nls}{\\\hspace*{1em}}

\newcommand{\arxiv}[1]{\href{https://arxiv.org/abs/#1}{\tt arXiv:#1}}

\title{Assurance of AI Systems\\[0.2ex]From a Dependability Perspective}
\author{\hspace*{-1.1ex}Robin Bloomfield (City, Univ.\ of London) and John Rushby (SRI)}
\begin{document}
\maketitle

\newpage
\vspace*{-0.6in}
\begin{center}
\textbf{Abstract}
\end{center}
\thispagestyle{empty}
\begin{quotation}

We outline the principles of classical assurance for computer-based
systems that pose significant risks.  We then consider application of
these principles to systems that employ Artificial Intelligence (AI)
and Machine Learning (ML).

On its own, testing is insufficient for assurance when very high
levels of confidence are required.  Hence, a key element in the
``dependability'' perspective is a requirement to have thorough
understanding of the internal design and operation (and hence
behavior) of critical system components and their interaction.  This
is considered infeasible for AI and ML because their internal
operation is developed experimentally over a limited (albeit large)
set of training examples and is opaque to detailed understanding.
Hence the dependability perspective, as we apply it here, aims to
minimize trust in AI and ML elements by using ``defense in depth''
with a hierarchy of less complex systems, some of which may be highly
assured conventionally engineered components, to ``guard'' them.  This
may be contrasted with what we call the ``trustworthiness''
perspective that seeks to apply assurance to the AI and ML elements
themselves by various forms of careful training, fine tuning, internal
``guardrails'' and automated examination.

In cyber-physical and many other systems, it is difficult to provide
guards that do not depend on AI and ML to perceive their environment
(e.g., other vehicles sharing the road with a self-driving car), so
both perspectives are needed and there is a continuum or spectrum
between them.  We focus on architectures toward the dependability end
of the continuum and invite others to consider additional points along
the spectrum.

For guards that require perception using AI and ML, we examine ways to
minimize the trust placed in these elements; they include diversity,
defense in depth, explanations, and micro-ODDs (Operational Design
Domains).  We also examine methods to enforce acceptable behavior,
given a model of the world.  These include classical cyber-physical
calculations and envelopes, and normative rules based on overarching
principles, constitutions, ethics, and reputation.

We apply our perspective to autonomous systems, AI systems for
specific functions, general-purpose AI such as Large Language
Models (LLMs), and Artificial General Intelligence (AGI), and we
propose current best practice and conclude with a fourfold agenda for
research in which we recommend development and application of: a) new
methods for hazard analysis suited to AI systems; b) layered
recursively structured architectures for runtime verification and
defense in depth; c) assurance for AI-based perception, and d)
improved understanding of human and machine cognition, shared
intentionality, and emergent behavior.

\memotwo{And diffusion models, reinforcement learning?}

\end{quotation}

\newpage

\thispagestyle{empty}
\tableofcontents

\clearpage
\clearpage
\setcounter{page}{1}

\renewcommand{\sectionmark}[1]{\markboth{\thesection.\ #1}{\thesection.\ #1}}
\renewcommand{\subsectionmark}[1]{\markright{\thesubsection.\ #1}}
\renewcommand{\headrulewidth}{0pt}
\setlength{\headheight}{13.6pt}

\section{Introduction}

\fancyhead[l]{\iffloatpage{}{\rightmark}}
\fancyhead[c]{}
\fancyhead[r]{}

Much discussion of potential risks with AI concerns ``existential''
threats \cite{AI-risk23}, but we suggest that lesser---yet significant
and widespread---hazards will arise as near-term AI is embedded in
other systems.  Hence, it is an urgent task to provide users of
systems incorporating AI, and also those responsible for their safety
and security, with methods to assess their hazards and to incorporate
appropriate means of mitigation, such as architectures that provide
assured runtime checking with diversity and defense in depth.
In addition, AI developers need to be aware of these issues and should
provide suitable APIs and other mechanisms that enable integration of
AI within assured systems.

By ``assured'' we mean that there are good reasons to believe that a
system will not exhibit certain harmful behaviors or will do so very
rarely, where ``rarely'' will be quantified as, say, no more than once
in a million demands.  ``Assurance'' is the process of justifying such
claims, so that it is rational to believe them and highly likely that
they will be borne out by subsequent experience.  Assurance is
normally required only for systems whose harms are such that they must
be rare.  Thus assurance is associated with methods for building high
quality or dependable systems and discussions of assured systems
generally (as here) have two interrelated threads: a) how to build
systems that are dependable, and b) how to justify confidence that
they are so.  

Risk owners in critical industries are responsible for understanding,
assessing and ensuring that hazards (i.e., circumstances with
unacceptable risk of leading to harm) are addressed according to
existing national and international frameworks.  Therefore, they need
methods for analyzing the potential harm (and benefit) from AI and
they need to develop justifications (e.g., as assurance cases
\cite{Bloomfield&Rushby:Assurance2}) for the safety and security of
systems that include AI---and their regulators will need to understand
and assess these.  We also note that unregulated industries still have
overarching responsibilities to address hazards and vulnerabilities.
Recently, many countries have established government agencies such as
the UK AISI (\url{aisi.gov.uk}) to support risk owners by developing
assessment methods and also by providing insights on AI systems that
may be components in these critical applications.  

There is much recent work to improve the reliability of AI-based
systems, and we welcome these developments.  However, although much of
this work has been successful, it does not amount to assurance, which
requires a rationally justified claim that failures will not exceed
some bound.  (Risk is generally understood as the product of severity
of a harm and its frequency.  If severity is assumed to be fixed, we
limit the risk by establishing a bound on its frequency, that is on
failures with respect to the assured claim.)  The contribution of
this report is to discuss the assurance challenges of AI and to
explain and explore classic strategies for achieving trustworthy
dependable systems, where \emph{dependable} is an umbrella term
encompassing safety, security, reliability, and so on.

The field of dependable systems engineering employs a standard
terminology \cite{Laprie:terminology}.  Specifically, \emph{failure}
occurs when a system departs from its explicitly specified or
implicitly expected behavior.  The cause of a failure is a
\emph{fault}, which may be the failure of some subsystem or component,
or an oversight or mistake in the design or implementation of the
system.  Faults precipitate \emph{error}s, which are incorrect values
or configurations of the system's internal state.  It may be possible
to detect and correct errors before they propagate to cause failure,
thereby providing \emph{fault tolerance}.  And failure need not always
be terminal: there may be some external processes for recovery or
adaptation that provide \emph{resilience} and allow (some aspects of)
the larger context to continue.

\begin{figure}[ht]
\centering{\includegraphics[width=0.7\textwidth]{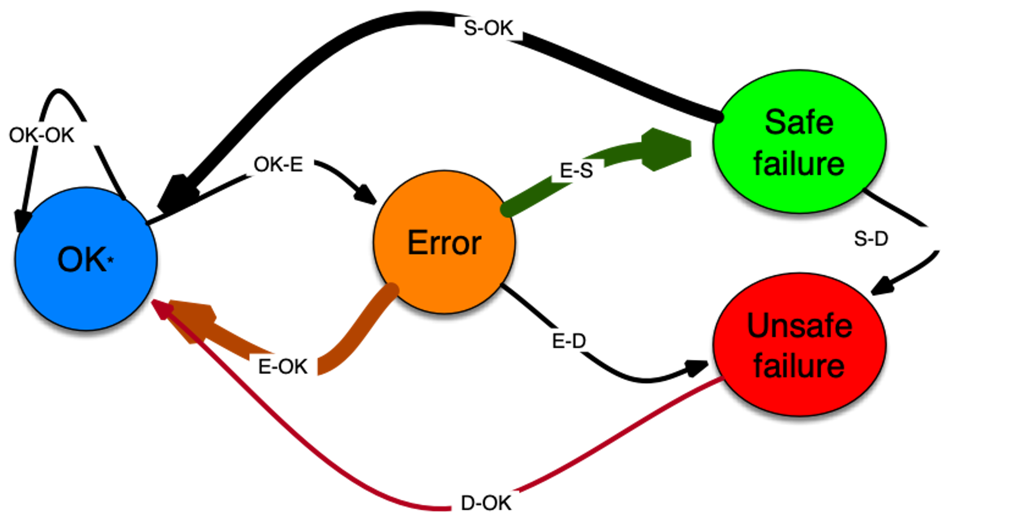}}
\caption{\label{four-state}Four State Model for Dependability}
\end{figure}

A 4-state model shown in Figure \ref{four-state} portrays transitions
on this ontology, corresponding to different circumstances and
strategies for dependable operation
\cite{ASCAD,Bloomfield&Rushby:FAISC24}.  These transitions illustrate
the three classic strategies for trustworthy dependable systems.

\begin{description}
\item[Fault avoidance.]  This aims to eliminate transitions from the
OK to Error state and is the domain of highly rigorous software
engineering.

\item[Fault tolerance.]  Faults correspond to transitions from the OK
to Error state, and Fault Tolerance to the transition back to OK.

\item[Failure Management and Resilience.]  Unsuccessful or missing
fault tolerance allows transitions from the Error to Failure states,
which we divide into Safe and Unsafe according to their consequences.
Failure Management tries to ensure the transitions are to Safe
Failures (e.g., a failed train comes to a halt) and provides some
recovery process that enables transition back to the OK state (e.g.,
the train is taken out of service).  There will be some loss
associated with the failure and recovery.

Resilience corresponds to the transition from Unsafe Failure back to
OK\@.  It may be associated with considerable loss due to both the
failure (e.g., a train crash) and the recovery or adaptation (e.g.,
revision to the signaling system).
\end{description}

\pagestyle{fancy}

These strategies are not mutually exclusive: dependable systems should
employ a combination, with emphasis varying according to the
information that is available, the nature of the technology being
used, and the consequences of failure.  For systems employing AI and
ML we assess the three strategies as follows.

\begin{description}
\item[Fault avoidance.]  As we describe in the following subsections,
this requires strong understanding of the system context and
requirements, and a means of assuring that these are applied and
implemented correctly.  Testing is an important means of assurance but
on its own it is insufficient for anything more than everyday risks.
We claim that the necessary understanding and assurance is currently
infeasible for AI and ML and consequently fault avoidance cannot be
the main dependability strategy for AI and ML systems.  Proposals for
``Guaranteed Safe AI'' do exist \cite{Dalrymple-etal24:GS-AI}, but we
consider them speculative.

\item[Fault tolerance.]  This is a key strategy when components are
relatively unreliable or hard to evaluate, but it does depend on
having well-defined safety/security requirements (so that we can
recognize transitions into the error state).  It is the main strategy
discussed in Sections \ref{ext} and \ref{specific}.

\item[Failure Management and Resilience.]  This becomes an important
strategy in circumstances where some failures are acceptable.  It is
particularly appropriate for new technologies or those used in ways
not previously envisaged, where some degree of learning by doing
is acceptable.  Resilience becomes a major strategy in Sections
\ref{generic} and \ref{agi}.

However, it is not directly applicable to systems that are too
important to fail, such as those where the risks are truly
existential, irreversible, or otherwise of high consequence to
society.  Nonetheless, the technology of resilience could be applied
to early warning and near-miss detection for these systems so that
effective controls can be imposed before they reach a critical stage
of development or deployment.
\end{description}

\newpage
\subsection{The Dependability Perspective on Assurance}
\label{trad}

Humankind has been concerned about the safety of their constructions
ever since they started making them.  From the beginning, they noted
failures, developed good practices, and specified liabilities and
penalties.  Nearly 4,000 years ago, the Code of Hammurabi stipulated:
\begin{quote}\raggedright ``If a builder build a house for some one,
and does not construct it properly, and the house which he built fall
in and kill its owner, then that builder shall be put to death''
\cite[Section 229]{Hammurabi}.
\end{quote}

Hazards to safety depend on what a constructed thing does and how it
does it.  Buildings, boats, bridges, and mines were among the earliest
constructions, and the hazards were that they would fall down, break
up, or catch fire, so nascent safety engineering and assurance focused
(not always successfully)
on ensuring that they were of adequate strength and were based on some
understanding of the mechanisms of stress and failure.  Active systems
such as boats not only needed to be strong, but to possess some form
of stability so that they would right themselves rather than tip over
in wind gusts, and a means of control so they could steer a desired
course.  And inherently dangerous constructions such as underground
mines would need to include escape routes from collapse or fire.
These concerns and methods were refined in the industrial revolution
as machines such as high pressure steam engines did new things and
introduced new hazards.  Systems such as railways introduced the need
for active procedures to ensure safe operation, such as signaling
protocols to prevent two trains using the same track.

Later, control systems became automated, first by mechanical systems
such as governors, then by analog electronic systems such as
autopilots, and then by digital computers.  Protocols and protection
systems also became automated, first with interlocks and then with
full automation implemented by digital computers.

Systems with control and procedural mechanisms implemented by
computers (so-called cyber-physical systems, CPS) drive the state of
the art in safety engineering and assurance today
\cite{Kopetz&Steiner22}.  The approach taken, which we call the
\emph{dependability} perspective, has fault avoidance as its primary
strategy and fault tolerance, generally for hardware failures, as its
secondary strategy, with resilience as a last resort.  A consequence
of this approach is that fault avoidance has to be assured to an
extremely high level: typically less than one failure in a billion
demands.  Testing is insufficient for this (see Section \ref{testing})
and must be supported by near-complete understanding of how the given
system works, what are its hazards, how these are eliminated or
mitigated, and how we can be sure all this is done and implemented
correctly.  The evidence and arguments that justify confidence in the
claims documenting this understanding constitute what is called an
\emph{assurance case} (see later for details and references).

So far we have considered systems where component faults are the main
cause of failures.  However, failures in complex systems are often due
to unanticipated interactions among components that are working
correctly (i.e., according to their requirements, which may themselves
by faulty).  These are called \emph{system failures} (as opposed to
\emph{component failures}) and many of the most egregious recent
failures are of this type (or a combination, where a component fault
escalates to catastrophe by precipitating a system failure).  Examples
include the nuclear accidents at Chernobyl, Three Mile Island, and
Fukushima; airplane crashes such as Air France 447 and the 737 MCAS;
spacecraft explosions such as Challenger and Ariane 5; marine
disasters such as Deep Water Horizon and the Norwegian frigate ``Helge
Ingstad''; financial failures such as Long Term Capital Management
(LTCM) and the banking collapse of 2008; miscarriages of justice such
as the UK Horizon scandal, and software outages such as SolarWinds and
CrowdStrike.  System failures were famously identified by Perrow as
``Normal Accidents'' \cite{Perrow84}.  A related notion, due to Per
Bak \cite{Bak13}, is ``Self-Organizing Criticality'' (SOC), which can
be seen as a ``hidden hand'' behind Normal Accidents
\cite{Lewis:resilience23}. 

System failures often result from drawing the system boundary too
narrowly: the system is considered to comprise the ``mechanism'' and
its immediate environment, including those who directly interact with
it, but the wider socio-technical context
\cite{Baxter&Sommerville:socio13} is overlooked.  Modern system safety
engineering and methods of hazard analysis such as Leveson's System
Theoretic Process Analysis (STPA) aim to identify and overcome these
wider sources of system failure
\cite{Leveson:STAMP,Leveson:saferworld16}.
An important consequence of AI is that it provides capabilities (such
as natural language) that cause it to become much more deeply embedded
in its socio-technical context than may be anticipated or recognized
and thereby extend the system boundary.
For example, Jatho and colleagues applied STPA to an ML-driven
prescription drug monitoring system and to a face recognition system
used in criminal justice and found previously unrecognized
socio-technical hazards in both cases \cite{Jatho-etal:ML-STPA23}.

We directed attention to the drawing of system boundaries when using
AI and ML in a recent paper \cite{Bloomfield&Rushby:FAISC24} and we
will return to this topic in Section \ref{agi}.  However, current AI
and ML systems are sufficiently unreliable that they are a significant
source of component failure in systems that employ them.
Consequently, we will focus on failures of AI and ML components,
but the larger context should be kept in mind.

Systems that use Artificial Intelligence (AI) and Machine Learning
(ML) are both an evolution and a step change from their predecessors:
although they often automate existing systems and procedures, they
work in different ways than what has gone before, and they can also do
intriguing new things.  Because they work in different ways, it is
difficult to apply established methods for safety engineering and
assurance to AI and ML, even when they are used in familiar or
slightly extended contexts, such as automated control and autonomous
systems.  In particular, the behavior of systems based on ML is
developed experimentally, so their inner workings are opaque and do
not support the understanding required for justified confidence.  And
because they can do new things, and appear to do them well enough to
substitute for humans in some circumstances, AI and ML are being used
in applications where they introduce entirely new hazards: by
substituting for people in activities that previously required human
levels of perception, language, intelligence, and judgment, failure
can go beyond physical harm and can affect personal wellbeing,
relationships, and society at large.

Beyond these fairly incremental progressions lies the step to
Artificial General Intelligence (AGI), with potentially superhuman
performance on significant activities, plus imagination, agency with
independent goals, and possibly consciousness.

Our aim in this report is to identify and briefly describe issues,
possible methods, and difficulties in assurance for systems with
significant AI and ML content.  We do this mainly from the
dependability perspective: because we do not believe that AI and
ML can provide well-assured fault avoidance, we accommodate them by
fault-tolerance strategies using \emph{guards} and/or \emph{diverse
replicas} or backups that monitor their behavior at runtime, often
within a larger architecture that provides \emph{defense in depth} and
resilience.
These architectures can deliver strong assurance for the overall
system with only weak assumptions on the behavior of AI and ML
components.  Furthermore, the concept can be replied recursively
within a system, so the AI and ML components can themselves have
internal ``guard rails,'' and also externally to the wider system and
organizational environment.

We are not the first to advocate these methods.  Notably, the interim
report from the Seoul Summit on Safety of Advanced AI identifies a
broad range of risks and issues and highlights the ``Swiss Cheese
Model'' of protection \cite[Section 5.1.2]{Seoul:interim23}, which can
be seen as a combination of diversity and defense in depth.

The dependability perspective may be contrasted with what we call the
\emph{trustworthiness} perspective, which does claim assurance (i.e.,
fault avoidance) for the behavior of AI and ML components that have
been developed, analyzed, tested, augmented, or restricted in various
ways \cite{Tenn-etal24:trustworthyAI}.\footnote{Terminology across
different fields is always difficult and sometimes contentious; the
field of dependability regards ``trustworthiness'' as a synonym for
``dependability'' \cite{Avizienis-etal04:dependability} and ``safety''
as a particular case within dependability, whereas AI tends to use
``safety'' as the generally required property and ``trustworthiness''
as the means for achieving and assuring it.  The adjective
``trustworthy'' carries a somewhat anthropomorphic tone that we
discuss in Section \ref{generic} and so we prefer the more neutral,
engineering terminology of dependability.  But then we need a name for
the alternative perspective and we use ``trustworthy'' for that
purpose.}  One criticism of the trustworthiness perspective is that it
often fails to assess potential harms realistically and the confidence
needed in their elimination.  Nonetheless, both perspectives have
merit when well executed and in practice there is a continuum or
spectrum between them.  In particular, we envisage guarded
architectures that are recursively structured where ``first level''
guards might use some AI and ML (e.g., for perception) and themselves
be guarded by simpler or diverse systems, eventually bottoming out on
conventionally engineered and assured guards so that the overall
architecture provides defense in depth.

The architecture and its assurance will vary according to how much
assurance ``credit'' is taken for trustworthiness of AI and ML
components \cite{Bloomfield-etal:disruptive19}.  We call this the
dependability/trustworthiness spectrum; a ``pure'' dependability
perspective takes no credit for trustworthiness of AI and ML
components, while a ``pure'' trustworthiness perspective claims full
assurance credit for those components.  Historically, the
dependability perspective, and its methods, are very similar to those
developed several years ago for using ``Commercial Off The Shelf''
(COTS) and ``Software Of Uncertain Pedigree'' (SOUP) components within
critical applications (e.g., non-ASIL software components in cars)
\cite{Bishop-etal03:COTS-SOUP,Profeta-etal96}.

In the remainder of this section, we describe the dependability
perspective on assurance for traditional systems that do not employ AI
or ML\@.  Subsequent sections introduce increasing amounts of AI and
ML and we discuss approaches and concerns regarding their assurance
from perspectives toward the dependability end of the spectrum.  We
invite others to provide complementary studies toward its
trustworthiness end.
We stress that the purpose of assurance as we
present it is not to impose a brake or burden on development, but to
support innovation by anticipating downstream hazards and suggesting
creative ways to mitigate them.

\subsection{Traditional Systems and their Assurance}

State of the art non-AI cyber-physical systems such as aircraft flight
control, safety systems such as nuclear shutdown, and all manner of
systems within critical infrastructure, medical devices, personal
gadgets and much else are generally engineered and assured for
suitably high levels of safety and other required attributes, such as
security or effectiveness, all generically referred to as
\emph{dependability} \cite{Evidence07}.\footnote{Strictly, security is
distinguished from dependability
\cite{Avizienis-etal04:dependability}: the former corresponds to the
impact of the environment on the system, whereas the latter is the
impact of the system on the environment.  For our purposes, we can
lump them together.}  In outline (a slightly more extended account is
provided at \cite{Bloomfield&Rushby:FAISC24}), the process for doing
this begins with identification of the potential hazards that the
proposed system might entail.  A hazard is a circumstance with an
unacceptably high risk of leading to harm or other undesired outcome
(the corresponding concept in security is \emph{threat}).  Hazard
analysis must consider more than the consequences of simple component
failures, it must consider malfunction and unintended function, and
also unexpected interactions among elements that are performing as
intended (recall the earlier discussion of system failures).  Methods
of hazard analysis often build on previous experience and may need to
be extended for new technology such as AI\@.  For example, HAZOP
\cite{McDermid96} uses \emph{guidewords} and asks ``what might happen
if this output is late/wrong/absent'' and so on.  AI may introduce new
kinds of error so that the guidewords may need to include phrases such
as ``is a lie,'' ``is biased'' or ``is offensive.''

With systems that do new things, or that operate in challenging
environments, there may be little relevant experience to guide hazard
analysis, so it is often supported by experiments (e.g., prototyping
or simulation).  In cars, for example, this is termed ``Safety of the
Intended Functionality'' (SOTIF) \cite{ISO-PAS-21448}; critics suggest
augmenting these (often massive, but still incomplete) experiments
with formal methods as these can, in principle, examine \emph{all}
cases within some context \cite{Saberi-etal:SOTIF20}.

Note that humans may be part of the system (e.g., as operators) and
their fallibilities and vulnerabilities must be taken into account.
Hazard analysis is conducted in the context of assumptions about the
environment in which the system will operate (which again may include
humans) and must consider (previously) unanticipated circumstances
within this context and also the suitability of the assumptions.
These are demanding tasks, and hazard analysis is not an exact
science: even its most effective methods can be imperfect and their
application requires skill, knowledge and experience
\cite{Leveson-etal10:STPA}.

As hazards are identified, the system and its evolving design are
adjusted to eliminate or mitigate them.  For example, if fire is a
hazard, we may try to eliminate it by removing sources of ignition and
fuel; if that is impossible or inadequate, we can try to mitigate the
hazard by adding a fire suppression system.  But then we have new
hazards concerning failure of that suppression system.  Note that we
usually try to separate those parts of the system concerned with
elimination and mitigation of hazards from those parts that deliver
its general functionality: the goal is to minimize the size and
complexity of those parts that need the highest levels of assurance.
We will also want to protect these critical parts from the rest of the
system: a practice known as \emph{partitioning}
\cite{Rushby99:partitioning}.  Of course, some aspects of the system's
general functionality may also be considered critical and they, too,
will be partitioned to the extent possible, and subject to assurance.
And some auxiliary functions such as logging may also be considered
critical as they will be needed to support forensic investigation in
the case of failure (consider the difficulty in conclusively
demonstrating failures of the British Post Office Horizon system
\cite{Christie:horizon20,Marshall-etal:recs21}).

After some rounds of iteration on hazard identification and
modifications to the system goals and design, we will have a set of
\emph{requirements} for the critical \emph{desired behavior} of the
computer control system that, with high confidence, ensures
dependability of the overall system and accurately characterizes its
assumed environment.  Identification and articulation of properties
assumed about the environment are fundamental to formulation of
requirements and are often the most difficult and fault-prone aspects
of the entire system engineering endeavor.  The analysis and reasoning
that shows that the requirements ensure safety and other critical
properties within their environment is an assurance task that we term
\emph{dependability requirements validation}.

Requirements concern what the system will do, not how it will do it,
so they should largely be described in terms of changes the system is
to bring about in the environment (this is a key insight due to
Michael Jackson \cite{Jackson95}).  How the system will do its task is
developed in \emph{specifications} for the \emph{defined behavior} of
the system and the \emph{architecture} of its components.
Architecture is a generalization of partitioning (often portrayed by
``boxes and arrows'' diagrams) and its purposes are to identify
\emph{fault containment regions} that limit \emph{fault propagation}
among components, to identify \emph{critical components} and limit
their complexity (because complexity is a source of faults and also
makes it more difficult to discover what faults may be present), and
generally organize things so that dependability relies on only the
architecture and the defined behavior of the critical components
\cite{Rushby-etal:DASC08}.

We then \emph{implement} the system according to its specifications
and architecture.  The mechanisms that ensure an architecture is
faithfully represented in the system implementation are among the
most difficult engineering challenges in computer science (involving
operating systems, ``buses,'' distributed consensus, state-machine
replication, transaction mechanisms etc.) and should employ only
well-attested techniques and products, with no ``homespun'' solutions
\cite{Rushby01:emsoft}.  During implementation, we may discover new
hazards and the whole process iterates: the new hazards cause
revision to the requirements\footnote{Confusingly, these revisions
are often called \emph{derived requirements} (the term comes from
avionics); it is confusing because their essence is that these
requirements were \emph{not} derived during the main process of
requirements development.} and their safety validation, and also to
the specifications and hence to the implementation.

Assurance is developed during and following this process.  After
dependability requirements validation, assurance divides into
three \textbf{verification} tasks.  (Verification differs from
validation in that, in principle, it can be performed with perfect
accuracy.)
\begin{description}
\item[Intent.] The specifications must be shown to be correct and
complete with respect to the requirements, subject to properties of
the architecture and assumptions about the environment

\item[Correctness.] The implementation must be shown to be correct and
complete with respect to the specifications, subject to properties of
the architecture and assumptions about the environment.

\item[Innocuity.] Any part of the implementation that is not derived
from the requirements must be shown to have no unacceptable
impact.\footnote{Software libraries provide an example of
implementation content that is not derived from requirements: we might
require only some trigonometric functions, but the whole library is
installed as part of the implementation.}
\end{description}

Different industries have their own standards and guidelines that
codify aspects of this process, often in great detail; the very
generic and abstract description given above is based on the
\emph{Overarching Properties} (OPs) proposed as the basis for future
civil aircraft certification in the USA \cite{Holloway:OP19}.\footnote{OP
concerns software assurance, so dependability/safety validation is
outside (occurs prior to) its scope.  Also, OP speaks of desired and
defined behavior rather than requirements and specifications.}

\begin{figure}[t]
\vspace*{-4ex}
\begin{center}
\includegraphics[width=1.0\textwidth]{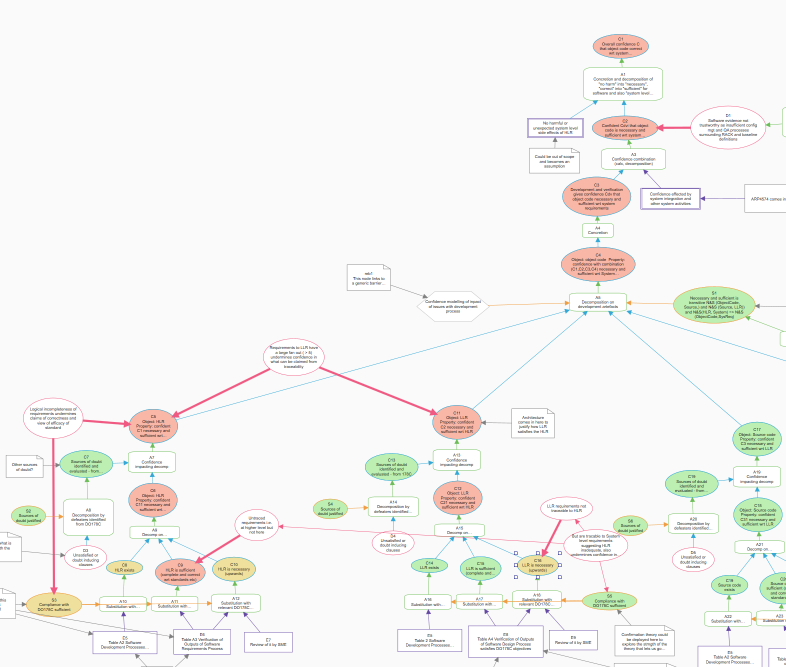}
\end{center}
\vspace*{-2ex}
\caption{\label{traffic}Portion of Graphical Rendering of an Assurance
Case Argument}
\vspace*{-2ex}
\end{figure}

Each of the assurance validation and verification tasks states that
some properties ``must be shown'' to hold; by this, we mean that there
must be reasons why the properties hold, and these reasons must be
clearly articulated and justified.  The state of the art for doing
this is an \emph{assurance case} (a generalization of safety cases
\cite{Kelly:GSN,ASCAD}) that provides an organized presentation based
on \emph{claims}, \emph{evidence}, and \emph{argument}
\cite{Bloomfield&Bishop:SSS10,Rushby:Cases15}.  Claims identify
properties of the system and/or its environment; evidence refers to
observations, measurements, or experiments on the system or its means
of construction or on its environment that justify certain claims; and
the argument uses the evidence to establish a hierarchy of claims
culminating in a significant \emph{top claim}.  The arguments of an
assurance case are not free form but \emph{structured} as hierarchy of
argument steps, each of which establishes a ``parent'' claim on the
basis of one or more ``child'' claims (we usually say
\emph{subclaims}) established at lower levels, or by evidence.  A
portion of a graphical rendering of an assurance argument is displayed
in Figure \ref{traffic}; our preferred treatment of modern assurance
cases, which we call Assurance 2.0, is presented elsewhere
\cite{Bloomfield&Rushby:Assurance2,Assurance2-home} and builds on the
ideas that a strong assurance case should be \emph{indefeasible}
\cite{Rushby:Shonan16,Bloomfield&Rushby:confidence22}, based on
established \emph{theories} \cite{Varadarajan-etal:DASC24}, and
subjected to \emph{dialectical examination}
\cite{Bloomfield-etal:defeaters24}. Barrett and colleagues describe
application of Assurance 2.0 to an AI safety case
\cite{Barrett-etal:confidence25}.

\subsection{From Assurance to Dependability}
\label{testing}

The focus on dependability validation and verification with overall
justification presented as an assurance case might seem like good
practice and a sensible way to develop high quality systems, but why
is it needed for assurance?  Why don't we just test the thing?
Indeed, whenever there is a major systems failure, the first reaction
of the press and public is ``they didn't test it enough.''  But in
fact, testing is insufficient and the reason is the extraordinary
levels of confidence required for safety and other critical properties
and, consequently, the infeasibly large number of tests that would be
required to validate them by observation alone.  We give a few numbers
for illustration.

In commercial airplanes, ``catastrophic failure conditions'' (those
``which would prevent continued safe flight and landing'') must be so
unlikely that they are ``not anticipated to occur during the total
operational life of all airplanes of a given type'' \cite[Section
3.2.4]{10-9-new}.  The ``total operational life of all airplanes of a
given type'' is about $10^{8}$ to $10^{9}$ flights for modern
airplanes.  With an average flight duration of about 90 minutes, this
requires a critical failure rate no worse than about $10^{-9}$ per
hour.\footnote{There are many measures for reliability, such as
failure rate, probability of failure on demand, probability of
fatality per mile, etc.  And the underlying system models may use
discrete or continuous time, with Bernoulli or Poisson failure
processes, etc.  The general conclusions drawn here are robust to all
these choices and so we do not describe them in detail.}

\phantomsection

Cars are among the most dangerous consumer goods with about 40,000
deaths per year in the United States and a fatal accident rate of a
little over 10 per billion miles.  It is intended that self-driving
cars should be safer than human drivers, so it might seem reasonable
(even though driving error is not the only cause of accidents) to set
the target at no more than 1 death per billion miles, which is around
$10^{-7}$ per hour, given an average speed of 30 mph.  However, that
is a rather technical assessment that pays no attention to likely
public reaction.  \label{driverless} Herbert Diess, the former CEO of
Volkswagen is quoted on their website with a more realistic
assessment: ``A ratio of ten-to-one is nowhere near good enough.  We
have approximately 3,200 traffic fatalities in Germany each year.  It
would be a disaster if we had even 320 deaths due to driverless
cars.''\exfootnote{In the USA, a ten-to-one ratio would yield 3,500
fatalities due to driverless cars, which would surely also be
unacceptable.}  Thus, it is plausible that the safety target for
self-driving cars should be 100 or even 1,000 times better than human
drivers,\footnote{\label{cruise}This sensitivity is illustrated by a
recent accident in San Francisco: a car with a human driver hit a
pedestrian and threw them under the wheels of a Cruise self-driving
taxi, which then dragged the victim for several
yards\excite{Cruise-accident24}.  Subsequently, Cruise lost their
license to operate in San Francisco (although their poor initial
response to the incident probably contributed to this harsh reaction)
\cite{Koopman24:cruise}.} which brings us into, or even beyond, the
requirements for commercial aircraft \cite{Koopman:howsafe22}.  But
commercial aircraft do get assured and certified and their safety
record justifies this, so why is $10^{-9}$ seen as such a challenge?

The answer is that we do not assure aircraft solely by testing.
If we test a system for $n$ hours and observe no failures, then in the
absence of other information, the best prediction we can make is that
the likelihood of no failures in another $n$ hours is about 50--50
\cite[p.\ 73, and sidebar on p.\ 74]{Littlewood&Strigini93}.  Hence,
to secure assurance for failure rates of $10^{-9}$ per hour we would
need to test the system for about $10^9$ hours, or around 115,000
years.  Even with 1,000 copies of the system on test, this is still
well over 100 years of continuous operation and is completely
infeasible
\cite{Butler&Finelli93,Littlewood&Strigini93,Kalra&Paddock16:miles}.

\label{stat-testing} It is not just the problems of developing
sufficient test data that makes this assurance route challenging and
often infeasible.  For modest values such as $10^{-3}$ it might seem
feasible to assess probability of failure directly by testing, but
ensuring statistical validity is a very demanding process where the
tests must be selected randomly from the \emph{operational profile}
\cite{NUREG-7234}.  It is demanding because the test harness must
reproduce the actual environment, the operational profile must be
accurate and must be sampled accurately (simply using the system
``randomly'' does not do this), the test oracle (which determines
success or failure of each test) must be correct, and the number of
tests must be large enough to deliver a statistically valid conclusion
(and there must be no failures among them---the number of tests needed
in the presence of failures is astronomical).  In addition,
statistical extrapolation rests on assumptions that the future will be
like the past.  Validating this requires analysis of all software and
system components, plus the test harness and its equipment, to show
there are no time dependent effects such as state accumulation,
security exposure and so on.

Thus, assurance by testing alone is impossible for critical systems:
the required number of tests is too large and the assumptions too
onerous.  Consequently, the idea behind classical methods of assurance
for critical systems is to develop justified confidence that the
system contains few or no faults.  We do this by examining the design
and construction of the system.  From confidence in absence of faults
we predict low probability of failure.  This prediction responds to
the ``absence of other information'' qualifier on the efficacy of
testing stated earlier.  As explained below, the combination of
confidence in the absence of faults and a feasible amount of testing
then provides system assurance.

It is fairly straightforward to reason from low likelihood of faults
to low probability of failure, but assurance does not tell us the
likelihood of faults: what it does is provide us with
\emph{confidence} that this likelihood is low.  Hence we need some
method for quantifying confidence that will support an inference to
probability of failure.

Confidence can be expressed as a subjective probability, so if we are
$95\%$ confident that traditional assurance works, that means we
estimate there is only 5\% chance that the assured system contains
faults.  We can now use testing to explore the existence of those
potential faults but, unlike the previous case, we know something
about the system so when we see $n$ hours with no failures, we can
conclude (by what is called Conservative Bayesian Inference, CBI) that
we are likely to see another $10n$ with no failures
\cite{Strigini&Povyakalo13,Zhao-etal:CBI20}.  This reduces the amount
of testing required and another idea reduces it still further.  This
is Bootstrapping \cite{Strigini-etal22:bootstrapping}: we need
assurance for $10^9$ hours, but this is over the lifetime of the
system.  When the system is first deployed, we might be satisfied to
know there will be no dependability failures in the first year, and we
will have only a few instances of the system operational in that time.
So we might need confidence for only, say, $10^5$ hours, and testing
for this, given prior assurance and CBI, requires only $10^4$ hours,
which is perfectly feasible.  After the first year, we might seek
confidence for the next year and for the larger number of systems now
installed, but we will have the operational experience of the first
year and that should be sufficient (given the tenfold multiplier of
CBI) to deliver the required confidence in safe operation, and so on
for subsequent years.
As with statistical testing, bootstrapping relies on arguing that
future behavior will be like the past, so that topics such as changes
in the environment and time dependent effects such as state
accumulation, software updates and security exposure all need
to be addressed.

Koopman criticizes bootstrapping \cite[Section 8.4]{Koopman:howsafe22}
when this is based solely on tests and accumulating experience, but
accepts that it is sound when, as here, it is founded on justified
prior confidence.  He is correct, however, that bootstrapping exposes
early adopters to greater risk: in the first year, we are confident
the probability of failure is sufficiently low that we will not see a
failure in that first year of operation, but in the second year we are
confident we will not see a failure in the next year of operation
(i.e., two years in total), and so the estimated probability of
failure must be lower in this period than the previous one since the
exposure is twice as long.  This is ethically sound, however, because
although the estimated probability of failure is greater in early
periods, it is always sufficient to provide adequate assurance of
safety.  In terms of the UK ALARP approach (``As Low As Reasonably
Practicable'') \cite{Redmill:ALARP10}, we show the initial risk is
tolerable and then use successful operating experience to further
reduce our risk estimates within the ALARP region.

\label{repairs}

Bootstrapping is based on failure-free operation.  But suppose
experience in operation does reveal a failure, hopefully not
catastrophic (in commercial aircraft, for example, it is required that
no catastrophic failure may be caused by a single fault).  Any such
failure reveals an unanticipated fault, arrival rate, circumstance, or
hazard, and these could be precursors to a catastrophic failure.
Consequently, commercial airplanes operate in a legal and ethical
framework where all incidents and accidents are promptly reported and
dispassionately investigated.  The FAA issues Airworthiness Directives
mandating workarounds or corrections to detected faults; in extreme
cases it may temporarily ground the fleet (as it did for the 737 MAX
in 2019--2020).  Bishop \cite{Bishop13} constructs a statistical model
for this scenario and shows that, under plausible assumptions,
detection and repair of faults significantly increases long run
safety, even if the fleet continues to operate after a fault has been
discovered, and even if repairs may be imperfect.  It follows that
there is much value in monitoring, analyzing and, if warranted,
correcting all non-trivial failures and their precursors.  We will see
in Section \ref{spis} that this process can be systematized with
``Safety Performance Indicators.''

We can now see that traditional approaches to assurance, such as
described for OP, give us justifiably strong confidence that the
system satisfies its critical properties; this can then be augmented
by testing and operational experience to deliver justified confidence
(via CBI, Bootstrapping, and a case that addresses their assumptions)
that the system satisfies its dependability goals, and this can be
reinforced by monitoring during operation.

We will sometimes use expressions such as \emph{weak}, \emph{modest},
or \emph{strong} assurance.  We intend these to be interpreted as
qualitative indications of the failure rates they are able to support.
We have already seen that rates on the order $10^{-9}$ are required
for fatal accidents in airplanes and cars, and we will say this
requires \emph{strong} assurance.  Some safety concerns about AGI
systems speak of \emph{existential} risk and we might say this should
require \emph{exceptional} assurance such as $10^{-12}$, while modest
and weak assurance might correspond to $10^{-6}$ and $10^{-3}$,
respectively.  These latter levels correspond to those conferred by
the lower levels of established standards such as DO-178C (DALs C and
D) \cite{DO178C}, IEC 61508 (SILs 1--3) \cite{IEC61508} etc.

All these allocations of risk and assurance are speculative: the
actual values would depend on the number of systems deployed, their
usage, and the nature of the harms they may cause.  One of the
conclusions of this report is that the risks from AI applications need
further analysis to place them in the risk spectrum so that we can
relate the AI assurance challenge to known approaches.

The roles of testing, analysis, and architecture change as the
criticality of the system increases.  At modest criticality,
operational experience and statistical testing can be combined with
modest analyses to justify assumptions and produce a plausible
assurance case.  As criticality increases, evidence from program
analysis evidence becomes more dominant---e.g., using formal methods
to verify absence of faults---with test and experience evidence
playing a supporting role in validating assumptions (e.g., of tool
efficacy).  If analysis is difficult or infeasible, as it is for many
AI systems, then the role of architectural mechanisms such as runtime
verification becomes more important

In general, assurance needs analysis and architectural mechanisms
combined with testing.  The combinations can be analyzed and justified
using CBI for high assurance, a related theory known as N/T (``N over
T'') \cite{Bishop&Bloomfield02:ISSRE,Bloomfield96:t-of-n} for modest
levels, and ``probability of perfection''
\cite{Littlewood&Rushby:TSE12} for architectural mechanisms.

In the following sections, we consider systems with AI and ML
components that do not lend themselves to traditional methods of
assurance and we explore how, and to what extent, dependability can
nonetheless be achieved and guaranteed.

\section{Assurance for Systems Extended with AI and ML}
\label{ext}

In this section, we focus on systems that do fairly traditional things
but are now extended with capabilities enabled by AI and ML\@.
Autonomous CPS such as self-driving cars are canonical examples.

For traditional assurance, there must be good reasons why we believe
the system achieves its dependability goals and those reasons are
documented and justified in its assurance case.  Systems that use AI
and ML pose challenges to this approach.  For example, \emph{symbolic}
AI, which is sometimes seen as an alternative to ML, uses automated
deduction (theorem proving) to derive conclusions from premises
composed of a set of axioms describing some aspect of the world plus
observations about the current state of the world.  It is possible to
guarantee validity of some methods of automated deduction (e.g., SAT
and SMT solvers with certificates \cite{Mebsout&Tinelli16:certs}), but
soundness depends on the choice of premises, which may be unvalidated
and derived from an informal or empirical model, and also on the
computational resources available (deduction generally requires
exponential time, or worse, so we may need to accept whatever partial
analysis can be accomplished in fixed time).  ``Expert systems'' were
a type of symbolic AI popular in the 1980s where deductive procedures
were applied to a collection of ``rules'' that axiomatized some
domain.  The concern, and one of the reasons for the demise of these
systems, was that individually reasonable rules could collectively be
inconsistent or incomplete, resulting in faulty output
\cite{Rushby88:SQA}.  Modern symbolic AI uses improved technology but
its challenge to assurance remains largely unchanged.

In a similar challenge to assurance, rather than performing actions
that are effective and safe for reasons that can be articulated and
verified, ML components operate by learning suitable behavior during a
period of training.  Training typically defines empirically effective
``weights'' in a \emph{deep neural network} (DNN); there will often be
millions, or even billions, of individually adjusted
weights.\footnote{There are other ML techniques, such as Support
Vector Machines \cite{Steinwart&Christmann:SVM08} but, despite
different mechanisms, these pose similar challenges to assurance as do
neural networks.  Similarly, the specific ML architecture in which
neural networks are employed, such as reinforcement and inverse
reinforcement learning, or large generative language and diffusion
models, has little impact on the fundamental difficulty of assuring
ML\@.}  The hope is that if the system works correctly on the training
examples, then it will work correctly on all similar examples, but
there are no strong methods to guarantee this.

A combination of symbolic AI with ML is a popular current approach
known as \emph{neurosymbolic} AI\@.  The strengths and weaknesses of
the two approaches seem to be complementary, but this does not assuage
their assurance problems.

Traditional assurance requires good understanding of how the system
works because tasks such as intent verification must show that certain
properties hold in \emph{all} circumstances.  This is infeasible for
most AI and ML components because we lack detailed understanding of
their operation\footnote{Recently, there has been progress in
associating learned ``concepts'' with specific clusters of artificial
neurons \cite{Templeton-etal:interp24}, but this is still some way
from the understanding required for assurance.} and without this,
testing is inadequate.  However, an alternative or constituent part of
the overall system assurance process can be to check that properties
hold in the circumstances actually encountered during operation.  This
is \emph{runtime verification} \cite{rv} or, more boldly \emph{runtime
certification} \cite{Rushby:RV08}, where components, often generically
referred to as \emph{guards} (or \emph{monitors}
\cite{Mehmed-etal:monitor20,Hawkins&Conmy:safecomp23}), are added to
the system to check its behavior against its required or specified
safety properties, or conservative simplifications of these.  If a
check fails, then the system must take some remedial action to
maintain or restore safety.  Runtime checking is not always possible
(e.g., for perception, where we have no independent knowledge of the
world) and both checking and remediation add complexity to the system
and may themselves introduce failures and hazards, so this approach
requires careful engineering \cite{Kopetz:architecture22}.
Nonetheless, a plausible approach is to guard suitable AI and ML
elements with conventionally engineered components that perform
runtime verification and can be assured in the conventional way.  This
approach is endorsed in some industry guidelines such as F3269-17 for
unmanned aircraft \cite{F3269-17}.  To be coherent, the overall
assurance case should include claims that are supported by subcases
for the runtime checking and remediation components and an argument
that integrates these with the case for the primary system.

\subsection{Safety Performance Indicators}
\label{spis}

Before examining the details of runtime verification, we discuss a
valuable secondary use of the same technology.  Runtime verification
checks critical claims in the overall assurance case at runtime and
intervenes if these are violated.  We can imagine checking lesser
claims that appear in subcases supporting critical claims as
indicators of potential violations of critical claims.  For example, a
critical claim might be that a self-driving car will not collide with
pedestrians.  A lesser, supporting claim might be that the car should
provide a one meter buffer between itself and pedestrians.  Violations
of the lesser claim suggest that things are not working as intended
and that in-service adjustments may be needed in the system and its
assurance case.

Runtime checks of lesser claims are called \emph{Safety Performance
Indicators} (SPIs); Johansson and Koopman
\cite{Johansson&Koopman22:SPI} define them as data-supported metrics
that provide a threshold on the validity of assurance claims.  SPIs
may be collected and evaluated per system (so excessive violations of
the one meter buffer may indicate that this car's sensors are faulty
or dirty) or over the whole population (so statistically excessive
violations of the one meter buffer may indicate that the relevant part
of the system and/or its safety case are deficient).  In some cases,
the system assurance case may have ``dynamic'' elements that reference
SPIs and adjust confidence in assurance at runtime
\cite{Denney-etal:dynamic-cases15}.  We do not consider this at odds
with the requirement that assurance cases should be indefeasible:
indefeasibility applies to the critical overall claims, while SPIs
measure thresholds on lesser claims.  If concern propagates from
lesser to critical claims, then it means that the assurance case is
deficient, not that the criterion for indefeasible soundness is
flawed.  In addition to SPIs that function as leading indicators,
warning of possible problems ahead, lagging indicators can also be
useful by providing feedback on how well things are going.  The UL4600
standard for evaluation of autonomous products has extensive
requirements for SPIs \cite[Section 16]{UL4600}\cite{Koopman:UL4600}.

\subsection{Runtime Verification}
\label{rv}

To investigate runtime verification for systems with AI and ML
components, we need some general understanding of the likely overall
system architecture and the properties that will be checked.  The
top-level structure of almost any system that employs AI or ML follows
from a single insight, which is that any entity that interacts
effectively with some aspect of the world must have a \emph{model} of
that aspect of the world \cite{Conant&Ashby70}.  In particular,
cyber-physical systems are always based on a model of the controlled
``plant'' and its environment \cite{Francis&Wonham76}.  The model can
be thought of as a simulation of relevant aspects of the world
that allows the system to predict its behavior and thereby choose
appropriate actions.

We find it useful to distinguish two aspects of world models.  A
\emph{local model} represents the configuration of the system's
``world'': that is, the state of all relevant objects and attributes
in its immediate environment, in sufficient detail and precision for
the system to perform its task.  The local model is typically built by
the system's sensors and associated subsystems.  Additionally, there
is a \emph{domain model} that represents how the elements in the local
model interact: that is, how the world ``works.''  When we say ``world
model'' or just ``model'' we mean the combination of local and domain
model.

For pure control systems, the domain model may be a collection of
differential equations and the local model will be the sampled values
of the variables over which it is defined.  For systems that involve
procedures the domain model may include state machines, and for CPS it
may be described by integrated formalisms such as timed and hybrid
automata.\footnote{``Model based system/software engineering'' (MBSE)
builds and experiments with explicit simulation models using tools
such as StateFlow/Simulink and derives the system requirements and
specification, and sometimes its implementation (and sometimes does
that automatically), from the simulation.  A criticism of this
approach is that the focus on simulation introduces implementation
concerns too early into the process (it is difficult to simulate
requirements stated as constraints unless theorem proving is used, so
specifications are often substituted for requirements) and thereby
compromises dependability requirements validation.}  For autonomous
systems, we take the example of a self-driving car, where the local
model concerns the immediate road layout with its obstructions,
traffic signs and signals, the locations of other road users,
pedestrians, and so on, and their inferred intent.  The domain model
describes how such objects move (or do not move), including physics
for speed, acceleration, braking, momentum, centripetal force, tire
friction, human reaction time and so on.  The domain model is used to
predict how the local model will evolve in time.  In a self-driving
car, it may be programmed by human designers but it can also be
developed by ML\@.  In later sections we will consider systems that
infer their own domain models or are constructed on foundations that
have learned them in training.  In those advanced systems, the domain
model may include human behavior and even societal topics such as
ethics.

Systems that employ AI and ML often use them in their
\emph{perception} (\emph{sub})\emph{system}, which is used to build
and maintain their local world model.  Optionally, AI and ML may also
be employed in the \emph{action} (\emph{sub})\emph{system}, which uses
both local and domain models to calculate and execute behavior that
will advance the system's progress toward its goal, while maintaining
dependability.\footnote{There is often substructure within the action
system: generally a \emph{decision} system that plans what to do
(e.g., ``move into the nearside lane between cars \texttt{a} and
\texttt{b}'') and an \emph{execution} system that manipulates the
controls (accelerator, brakes, steering, horn) to perform that plan.
These details are unimportant at our level of discussion.}

It is often possible to guard AI or ML-generated actions with highly
assured conventional software that uses an explicit human-developed
domain model to check their safety against the local model.  However,
if the local model is constructed by a perception system that uses AI
or ML then we have to ask how its own accuracy can be assured or
guarded: it does no good to run safety checks against a faulty model.
However, in some cases it is possible for the guard to use a local
model that can be assured.

We can therefore distinguish two classes of guarded AI-enabled systems
according to the nature of their runtime verification.
\emph{Assuredly guarded} systems are those whose safety and other
critical properties can be checked and enforced by assured guards that
do not themselves use AI or ML, neither for perception nor action.  In
cases where the guards are so strongly assured that they can be
considered ``possibly perfect'' (or ``probably fault-free''), it is
possible to make very strong claims for dependability of the overall
system \cite{Littlewood&Rushby:TSE12,Littlewood-etal:RESS17}.  The
second class is \emph{unassuredly} guarded systems, where the guards
themselves depend on AI or ML, typically in their perception systems.
See, for example, the architectures listed on page \pageref{aiarchs},
where those that include a traditionally assured backup guard are
assuredly guarded and others (except number 1, which is not guarded)
are unassuredly guarded.\exfootnote{We use ``assurable'' to refer to
components (such as traditionally engineered guards) that can be
assured in isolation, and ``verifiable'' to refer to those that can be
assured in their system context (e.g., assurable guards that do not
depend on an ML-generated model).}

An example of the first class is an unmanned aircraft controlled by AI
and ML that is deemed safe as long as it remains within some specified
portion of airspace (a ``geofence'').  This is a constraint that can
be checked by a conventional navigation system and enforced using a
conventional guidance system as an override when violations are
detected \cite{Hayhurst16:safeguard}.  Here, the navigation system
provides a local model for the guard and its guidance override system
uses a human-constructed domain model.  Many robots can be guarded by
``virtual cages'' of this kind.  Notice also that when formulating the
initial ``concept of operation'' for a system, it may be possible to
adjust the concept so that assurable guards become feasible.  For
example, rather than an autonomous shuttle bus sharing its route with
other vehicles and pedestrians, it could use a dedicated and isolated
track, and would not then require a sophisticated AI-based perception
system.

An example of the second class is a self-driving car where the action
guard uses an assurable human-provided domain model to check
AI-generated actions for safety and other dependability properties
against a local model of its environment---i.e., location of other
road users and pedestrians, interpretation of traffic signs and
markings, detection of road layout, etc.  But to fully verify the
guard we need assurance for accuracy of its local model, and to do
this in the traditional way we would want to see a strong argument
that the model satisfies its dependability requirements and that these
are sufficient to mitigate its hazards.  As we examine in the next
section, this is typically infeasible for models constructed by an AI
perception system that uses ML to interpret its sensors (although
recent developments are starting to explore some of these topics
\cite{Hu-etal20:perception-rqts,Celik-etal:STPA-perception22,Bloomfield:templates-autonomy21}).

\subsection{The Challenge of Assuring Perception}
\label{perception}

We have seen that an action subsystem can generally be assuredly
guarded, but is dependent on a model of the local world constructed by
a perception subsystem that is harder to guard.

A possible compromise is for the action guard to use a simpler,
conservative local model constructed by conventional and highly
assured software.  This is sometimes seen in self-driving cars where
the guard is a simple system for automated collision detection and
emergency braking, similar to the Automated Driver Assistance System
(ADAS) that is provided for some human-driven cars (some of these
themselves use ML-based perception---we consider this case
later).\footnote{The US National Highway Traffic Safety Administration
recently finalized a rule (FMVSS No. 127) that will require Automatic
Emergency Braking (AEB), to be standard on all passenger cars and
light trucks by September 2029.  Tests by the American Automobile
Association found that current AEBs (which are built to a lower,
voluntary standard) are somewhat effective against rear-end
collisions, but not at all effective against sideways collisions at
intersections.}  However, this does not provide full assurance because
forward collisions are not the only hazards (Mehmed and colleagues
cite data from a NHTSA study of 5.9 million human-driver accidents
that used a classification with 37 categories
\cite{Mehmed-etal:eval22}), nor are human accidents necessarily good
models for failures of an autonomous system, and nor is emergency
braking attractive as the sole method of hazard mitigation.  And
notice that to prevent excessive activation of the emergency system,
the primary system of a self-driving car must take its behavior and
capabilities into account.  For example, the primary action subsystem
may calculate (correctly) that a certain maneuver is optimal and safe,
but that it is also likely to activate the more hair-trigger response
of the emergency system, and so it must choose a different, suboptimal
behavior.  In general, systems employing defense in depth must be
designed and developed as a whole to ensure that unintended
differences among the layers do not cause unnecessary loss of
availability, while at the same time striving to preserve safety
arguments based on diversity.  These are challenging requirements.

Furthermore, accidents and collisions are not the only hazards that
should be considered: for example, the City of San Francisco has
reported dozens of incidents where robotaxis interfered with emergency
responders, and there must be many other circumstances where
self-driving cars increase risk or inconvenience for others without
themselves being involved in a collision
\cite{Koopman&Widen:redefining24}.

For verified dependability, assurable guards must detect all hazards
and mitigate them safely, without excessive false alarms.  It is
possible that a more comprehensive suite of verifiable ADAS-like guard
functions could do better than a single safety system, but they would
have to steer a difficult path among incomplete coverage, false
alarms, and unattractive, abrupt, responses.\exfootnote{ADAS is very
effective for human drivers, but the failures of human drivers may be
different to those of self-driving cars.}  A contrary point of view is
that although there may be many circumstances leading to accidents,
the exact circumstances are irrelevant to ``last second'' detection as
there are only a few possible emergency responses: essentially,
braking or evasive action, and so an ADAS-like guard (or a suite of
such guards) could be an acceptable means of assurance, provided we
can develop assurance that it will always select and perform
appropriate emergency responses (see Section \ref{odds}).

An argument against simple ADAS-like guards is that more sophisticated
perception could detect hazardous situations earlier and more
completely, and provide less abrupt mitigation.  Accordingly, it is
worth considering guards that do use AI and ML and asking whether they
can be assured for trustworthiness within an overall approach that
remains close to the dependability end of the
dependability/trustworthiness spectrum.  For example, we could imagine
an ``ML-friendly'' adjustment to traditional assurance where we do
construct requirements and identify hazards, and then develop training
data of ``sufficient'' size, coverage, and quality\footnote{ML
requires such a vast quantity of training data that the training sets
are often labeled automatically, by another AI system, and therefore
cannot be considered high quality.} to encompass all of these and use
it with a well-regarded ML toolset to generate the perception
capabilities desired.  In fact, this is exactly what is proposed by
several groups engaged in research and development of autonomous
systems
\cite{Dong-etal:MLassurance24,Bensalem:LECS23,Anderson&Dennis:Autonomous-Nuclear23},
although they generally consider primary systems rather than guards,
and only one focuses specifically on perception
\cite{Salay-etal:perception22}.

The hazard most generally recognized in perception using ML is lack of
``robustness,'' meaning that small changes in sensor data may cause
its ML-generated interpretation to change abruptly.  This concern is
validated by so-called ``adversarial examples''
\cite{Szegedy-etal:adversarial13}.  Typically demonstrated on image
classifiers, the examples are deliberately constructed minor
modifications to an input image that are indiscernible to a human
observer but cause an ML classifier to change its output, often
drastically and inappropriately.  Image masks have been developed that
will cause misclassification when overlayed on any image input to a
given classifier, and there are universal examples that will disrupt
any classifier \cite{Moosavi2017universal}.  Furthermore, there are
patterns that can be applied to real-world artifacts (e.g., small
images that can be stuck to traffic signs) that will cause them to be
misread by an image classifier \cite{Brown-etal:patch17}.

There is much work on detection and defense against adversarial
attacks (see \cite{Huang-etal:survey20} for a survey) and on the
threats posed by general lack of robustness
\cite{Vassilev-etal:adversarial24,Kumar-etal:adversarial20}.  A
problem with all this work is that the techniques for guaranteed
robustness have so far scaled only to relatively simple systems and
not, for example, to the object detector of a self-driving car.  And,
more importantly, robustness is not the topic we really care about: we
want assurance of accurate perception.
There is work that verifies contracts on some aspects of perception,
notably detection of traffic lanes
\cite{Astorga-etal:perception-verif23}, but this particular problem
can also be solved without ML \cite{Duda&Hart72:Hough} (although those
solutions may also be hard to verify).

Accurate perception for world models in general depends not only on
robust interpretation of individual sensors but also on effective
sensor fusion.  This is illustrated by the fatal accident between an
Uber self-driving car and a pedestrian walking a bicycle in Arizona on
18th March 2018 \cite{NTSB:Uber18}.  The Uber car used three sensor
systems (cameras, radars, and lidar) and fused them using a priority
scheme that delivered a ``flickering'' identification of the victim as
the sensor systems' own classifiers changed their identifications, and
as fusion preferred first one sensor system, then another, as listed
below \cite[Table 1]{NTSB:Uber18}.

\newpage
\begin{itemize}
\itemsep=0.6ex
\item
    5.6 seconds before impact, victim classified as \emph{vehicle}, by radar
\item
    5.2 seconds before impact, victim classified as \emph{other}, by lidar
\item
    4.2 seconds before impact, victim classified as \emph{vehicle}, by lidar
\item
    Between 3.8 and 2.7 seconds before impact,\nls classification
    alternated between \emph{vehicle} and \emph{other}, by lidar
\item
    2.6 seconds before impact, victim classified as \emph{bicycle}, by lidar
\item
    1.5 seconds before impact, victim classified as \emph{unknown}, by lidar
\item
    1.2 seconds before impact, victim classified as \emph{bicycle}, by lidar
\end{itemize}
Consequently, the object tracker never established a trajectory for
the victim and the vehicle collided with her even though she had been
detected in some form or other for several seconds.  The car in this
incident used a particularly poor method of sensor fusion and better
methods are now employed, but it is not known how to provide assurance
for their behavior.

We conclude that approaches based on trustworthy perception do not yet
provide an adequate basis for assured local models, although they do
represent good practice and can be welcomed and used on that basis.

We are therefore torn between guards that do not use AI or ML and are
potentially assurable, but whose interventions are crude and possibly
unacceptable, and those that may be acceptable but are unassurable
because they use AI and ML for perception.  However, there is one last
approach that might provide hope: this is an argument based on
diversity and defense in depth.

\subsection{Assurance through Diversity and Defense in Depth}
\label{diversity}

In our discussion so far, we have been using guards for runtime
verification, where overall assurance depends on that of the guard.
But we could also argue that any guard, even if it is not assured,
will provide redundancy, and its development and implementation can be
completely independent and ``diverse'' from the primary system.
Hence, it might be argued that failures of the guard will be
independent of those of the primary and the combination could deliver
a multiplicative improvement in dependability (e.g., simplistically
ignoring issues of independence, two systems with $\pfd \leq 10^{-4}$
give us $10^{-8}$ overall).  We can also imagine a more integrated
system where, rather than a primary system and a guard, we have
redundant, diverse perception systems with different architectures and
training sets contributing to a single consensus model, or to one, but
still fused, model for operation and another more conservative one for
runtime verification of actions.

The topic of assurance through diversity is large and somewhat
contentious.  There is little doubt that architectures employing
diverse components are generally more reliable than single threads.
In particular, there is evidence that ``portfolio'' or ``ensemble''
perception systems are more reliable than their individual
constituents \cite{Ju-etal:ensembles18}.  The difficulty is in
demonstrating that diversity provides benefit in any \emph{particular}
case, and in estimating \emph{how much} benefit it provides
\cite{Littlewood&Miller89}.  In particular, there is no feasible way
to validate failure independence (it is a variant on the infeasibility
of assurance by testing),\footnote{However, it is feasible to validate
modest confidence in independence, and to detect indications of its
absence.} nor strong reasons for believing it.  This is because some
circumstances are just plain hard to interpret and it is possible that
all components may then fail together: consider the scenario with the
Cruise self-driving taxi described in Footnote \ref{cruise}---would
any test set have included pedestrians being thrown under the
wheels?\footnote{We would expect training to include other objects
(e.g., traffic cones) being thrown under the wheels, and the hard
common perception problem is then one of generalization.}
Furthermore, these difficult cases do not ``thin out'' as more are
considered: the distributions seem to have ``fat tails''
\cite{Koopman:heavy-tails18} (see footnote \ref{fattails} for an
illustrative example).  Thus, the ``multiplicative'' argument
for assurance by diversity is indeed simplistic.

However, although diversity alone cannot provide strong assurance, it
can provide a useful step in a ``ladder'' of assurance.  Nuclear power
generation usually employs such a ladder of protection systems
providing \emph{defense in depth}: there is the operational control
system, designed to manage the plant efficiently and safely, then a
(safety) limitation system that can intervene to ensure the plant
behavior stays within some safe envelope but remains operational, and
finally a shutdown system that functions as an assured guard that
guarantees to initiate a safe shutdown when safety parameters are
violated.  The operational and limitation systems are carefully
engineered but do not guarantee the dependability goals established
for nuclear power: that is accomplished by the assured shutdown
system.  But the operational and limitation systems are diverse in
function and construction and although this does not provide a strong
basis for assurance, the presence of the limitation system almost
certainly reduces demands on the shutdown system.  This is beneficial
because an emergency shutdown is disruptive and expensive.\footnote{A
complete shutdown can be a complex operation involving both automated
initiation and intervention by expert operators, depending on the
state of the plant and reasons underlying the need to shutdown (cf.\
the Fukushima incident).}

The control, limitation, and shutdown systems in this architecture for
nuclear power generation all use traditional software, but a similar
approach could be used in systems with AI and ML\@.  For example, in a
self-driving car the full functionality could be provided by a primary
system employing AI and ML that is supported by a diverse system, also
employing AI and ML, that is focused on safety, with assurance
provided by traditionally engineered ADAS-like emergency backup
functions.  The diverse primary and safety systems deliver acceptable
behavior and reduce demands on the assured backup so that its
interventions, though crude, are rare and tolerable.  Example AI
safety functions include ``Responsibility-Sensitive Safety'' (RSS)
\cite{Shalev-etal17} and an ``AI Safety Force Field''
\cite{Nister-etal:SFF19} that avoid creation of unsafe situations, and
``REDriver'' \cite{Sun-etal:REDriver24}, which monitors proposed
trajectories of self-driving cars against Chinese traffic laws.

A criticism of these particular proposals for defense in depth is that
the diverse safety-focused system is concerned solely with actions and
relies on the same local model as the primary system.  If the
perception system fails to detect a pedestrian, for example, then it
will be absent from the common model and the safety-focused system can
deliver no protection: everything will depend on the emergency backup.

One possible mitigation for this hazard is to provide the
safety-focused system with separate sensors and perception so that it
can build its own local model; alternatively, it could use the same
sensors as the primary system but with diverse perception software
where the local model for the safety system could be simpler than that
for the primary system.  For example, the primary system in a
self-driving car must not only detect objects, but their type (car,
bicycle, pedestrian etc.) and fine details such as where drivers and
pedestrians are looking, because the action system needs to project
their likely motion and future position, whereas the safety system
could just perceive undifferentiated objects and surround them with a
conservative ``safety box.''  Intuitively, we might think that the
simpler model of the safety system could be more trustworthy, but it
is challenging to provide credible assurance for this.  Nonetheless,
diverse sensors and/or diverse perception can provide the primary and
safety-focused systems with different local models (e.g.,
\cite{Hanselaar-etal:SafetyShell23}) and this should improve safety,
but it is difficult to manage diverse models without false alarms
(although these can be verifiably avoided in some circumstances
\cite{Mehmed-etal:formal-falsepos20}).  

Rather than manage two models,
it seems preferable to fuse the products of the diverse perception
systems into a single local model that combines precision and safety
but, as illustrated by the Uber crash described earlier, fusing can
introduce flaws of its own.  A more attractive approach is to
construct a single local model using diverse perception systems in a
principled way.  Conventional perception systems work ``bottom up'':
one or more deep neural nets take sensor data (e.g., images from
cameras or point clouds from lidars) and deliver interpretations
(e.g., lists of detected objects) that are further processed and fused
to produce the local world model.  One argument against this approach
is that it works ``backwards'' from effect (image) to cause (objects),
which is inherently difficult.  Another is that it prioritizes
fleeting sensor data above the local model, which is the repository of
much accumulated information.  An alternative approach, and the way
human perception is believed to work \cite{Bennett23:history}, reasons
``forwards'' using the model to \emph{predict} sensor data (or basic
interpretations thereof) and then applies a form of Bayesian inference
known as \emph{Variational Bayes} \cite{Kingma&Welling13} to optimize
the model in a way that minimizes prediction error.  Notice that
prediction errors provide continuous feedback on accuracy of the local
model.  Furthermore, minimization of prediction error provides a
principled way to fuse diverse lower-level sensor and perception
functions.

In humans, this mechanism for perception is known as \emph{Predictive
Processing} (PP) \cite{Wiese&Metzinger:VanillaPP2017} and it is
believed to be coupled with a \emph{dual-process} architecture
\cite{Frankish10}.  The lower-level process, known as ``System 1''
\cite{Kahneman11}, performs rapid unconscious perception using PP so
long as prediction errors are fairly small, indicating the world is
evolving as expected.  A large prediction error is called a
``surprise'' and the higher-level ``System 2'' process intervenes to
resolve it using more deliberative cognition and a more comprehensive
domain model.  For example, a self-driving car using a perception
system of this kind may find that a vehicle some distance ahead that
has been seen and correctly predicted for some time disappears,
provoking a surprise; System 2 may hypothesize that the disappearance
is due to it being occluded by another vehicle that has not been
detected (e.g., a white truck against clouds); System 2 can then add
the hypothesized occluding vehicle to the local model and it will be
included in future predictions, thereby sensitizing basic sensor
interpretation to look for it, and the action system can also initiate
cautious defensive actions such as a lane change.\footnote{This
scenario is motivated by a Tesla crash in Taiwan on 1 June 2020:
\url{https://www.youtube.com/watch?v=ZmHBA_vV39w}.}  System 2 in this
dual-process architecture is a location for diverse interpretation of
the local model (and updates to resolve surprises) and also,
optionally, the guarding of actions
\cite{Jha-etal:Safecomp20,Salay&Czarnecki:perception22}.

A dual-process architecture with System 1 using PP to integrate
diverse perception systems and with System 2 providing diverse
perception refinement and action guards is an attractive arrangement
for autonomous systems.  Although assurance for its AI and ML
components will be weak, its overall architecture has a rational
structure that can justify modest confidence in assurance based on
redundancy and diversity.  Modest confidence is insufficient for
system dependability, but that is not the claim it needs to support:
instead, contributing to defense in depth, it supports the claim that
it reduces demands on the traditionally engineered emergency backup to
a tolerable level,\footnote{For example, the primary system of a
self-driving car might see an object ahead and classify it as a
cardboard box and be prepared to drive over it; as it comes closer,
the emergency backup will see it as an unclassified obstruction and
slam the brakes on; but the safety system will also have seen it as an
unclassified obstruction and could have changed lanes to avoid it
``just in case.''  On a larger scale, the safety system might veto a
time-optimal route selection by the primary system because it
considers it unduly hazardous (e.g., potentially icy).} and
dependability is assured by the backup (the overall assurance case
will integrate all these elements in its argument).

Even when a cyber-physical system does not employ a dual-process
architecture, a second layer of runtime verification can have general
benefit.  The primary runtime verification layer checks for immediate
safety hazards and intervenes to avoid or mitigate them, while the
second layer checks for SPIs that act as ``canaries in the mine'' and
warn of incipient or developing risks.

We have discussed several options for assurance of systems that use AI
and ML for autonomy or other advanced functions, so in the following
section we provide an enumeration and summary of some of the
architectures considered.

\subsection{Summary of Architectural Choices}
\label{archs}

We are concerned with assurance of cyber-physical and similar systems
that use AI and ML\@.  We recognized a spectrum of approaches ranging
from those that aim to develop trustworthy AI and ML components to
those that place no trust in those elements and instead guard them
with components whose dependability is achieved and assured by
conventional methods.  

We divide the critical AI and ML components into a perception
subsystem that builds a model of the local world and an action
subsystem that uses the local model and its own domain model to plan
and execute safe and effective actions.

Before proceeding, we explain the intended interpretation of the
diagrams that follow.  An arrow from one box to another indicates flow
of data, whereas an arrow from a box to a line (see examples below)
indicates the ability to intervene or override the flow of data on
that line.  The incoming arrows on the left indicate data sensed from
the environment, while the outgoing arrows on the right indicate
control data sent to the actuators.

\phantomsection
\label{aiarchs}
\begin{enumerate}

\item \label{bare} We begin at a point on the assurance spectrum where
the AI and ML are considered ``trustworthy.''  This does not amount to
credible assurance from the dependability perspective so both the
world model and the action subsystem are considered
unassured.\footnote{Those who wish to grant some confidence in the
trustworthiness of well-examined AI and ML components may read our
``unassured'' category as ``somewhat assured.''  ``Assured'' could
then be read as ``strongly assured,'' with ``weakly assured'' as an
intermediate category, where the adjectives indicate confidence in the
properties claimed for the components or system.  Further variants
would be those where the action subsystem uses no AI and is
potentially assurable.  Note that ``assurable'' concerns the ability
to \emph{guarantee} certain properties; some unassurable architectures
may deliver generally good behavior, but we cannot guarantee it.}
Hence, we next focus on the dependability end of the spectrum.\\

\includegraphics[width=0.9\textwidth]{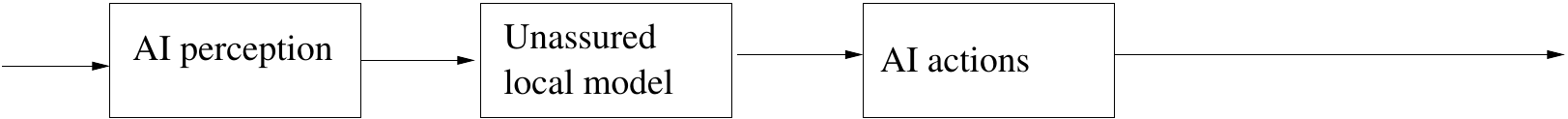}

\item \label{backup} As above with the addition of a conventionally
engineered and assured guard or backup.  The overall architecture is
assurable and is viable for applications such as a geofenced vehicle.
However, the coverage of the guard and the effectiveness and
acceptability of its interventions need careful justification for
applications such as a self-driving car where only crude
interpretations of the world can be constructed without ML.

\includegraphics[width=0.9\textwidth]{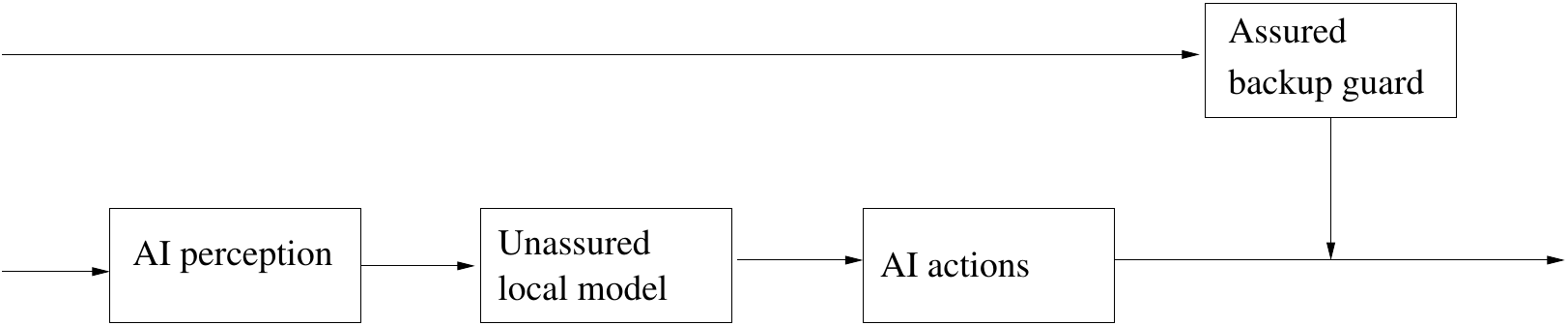}

\item \label{actionguard} As \ref{bare} but with a conventionally
engineered and assurable guard for the action subsystem, driven from
the same model as the primary system.  This is vulnerable to errors in
perception leading to a flawed world model and so the guard 
may be verified but it is not assurable and neither is the overall system.\\

\includegraphics[width=0.9\textwidth]{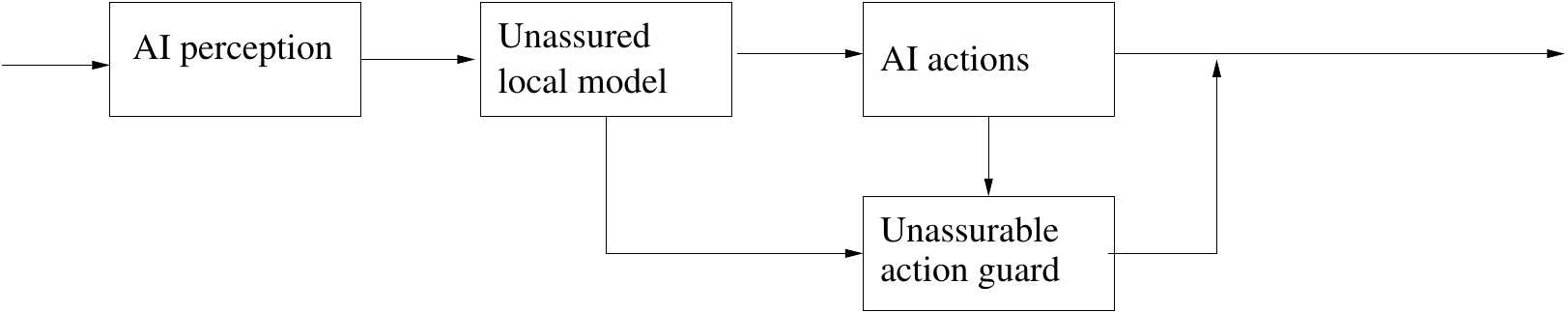}

\item \label{combo} Combination of \ref{backup} and \ref{actionguard}.
Overall, this is assurable due to the backup guard.  The action guard
is not assurable as it is driven by the unassured world model.
However, it should reduce demands on the backup, thereby improving on
\ref{backup}.

\mbox{}\\
\includegraphics[width=0.9\textwidth]{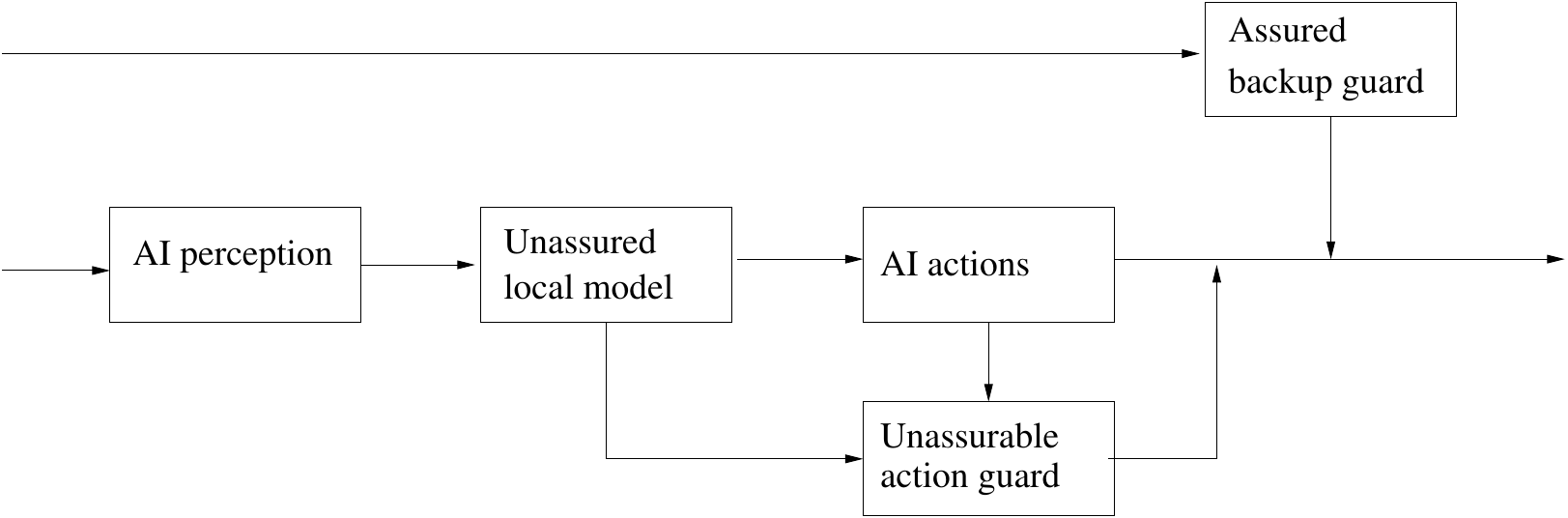}

\item \label{dualmodels} As \ref{actionguard} but with diverse
perception subsystems driving separate world models for the primary
action subsystem and its guard.  Diverse perception systems could
provide benefit but without some cross-comparison the local models are
not assured and neither is the guard and so the overall architecture
reduces to \ref{actionguard} and is not assurable.\\

\includegraphics[width=0.9\textwidth]{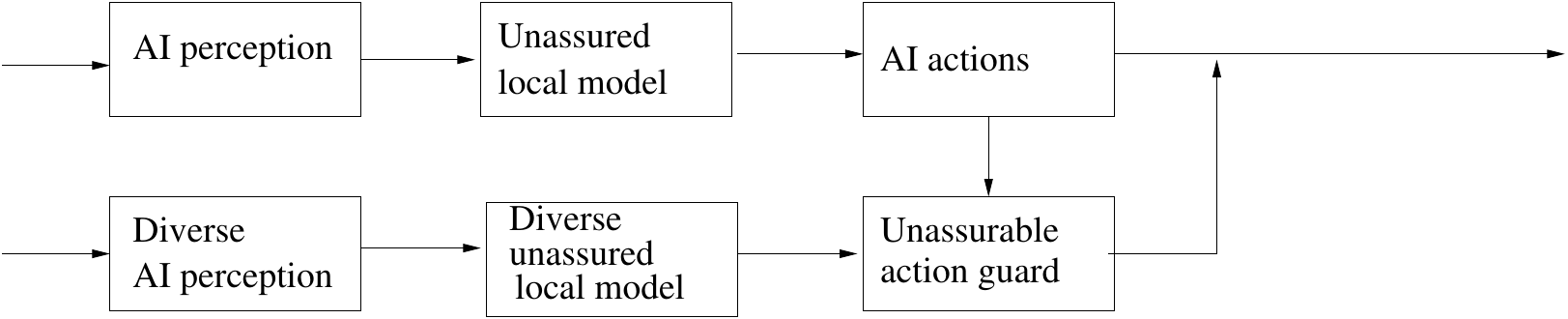}

\item \label{diversemodels} As \ref{actionguard} but with diverse
perception systems contributing to a single world model that is
considered weakly assured by a diversity argument.  This architecture
is weakly assurable overall: the traditionally engineered guard
provides assurance on actions but depends on a world model that is
only weakly assured.

\includegraphics[width=0.9\textwidth]{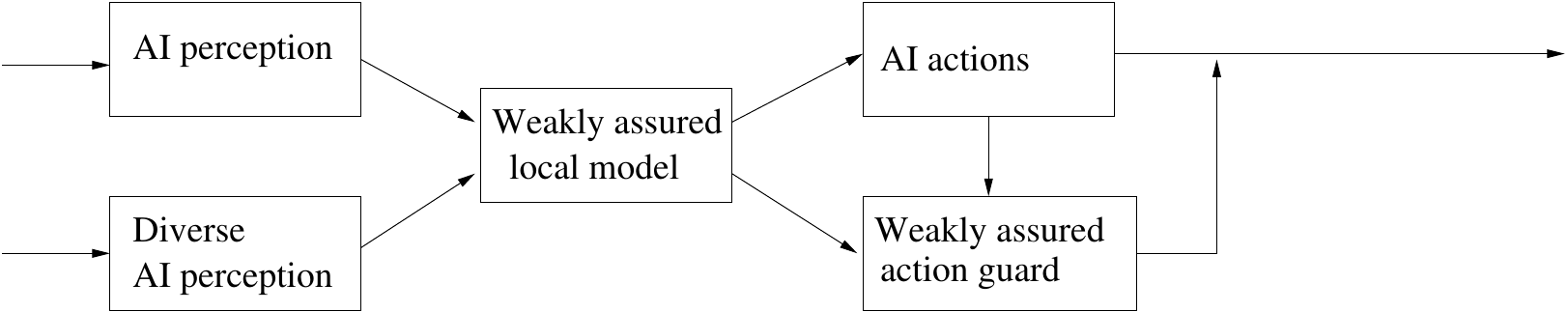}

\item \label{allin} Combination of \ref{backup} and
\ref{diversemodels}.  This architecture is assured by the assured
backup guard and improves on \ref{combo} because the architecture of
\ref{diversemodels} provides some assurance for reduction of demands
on the backup.

\includegraphics[width=0.9\textwidth]{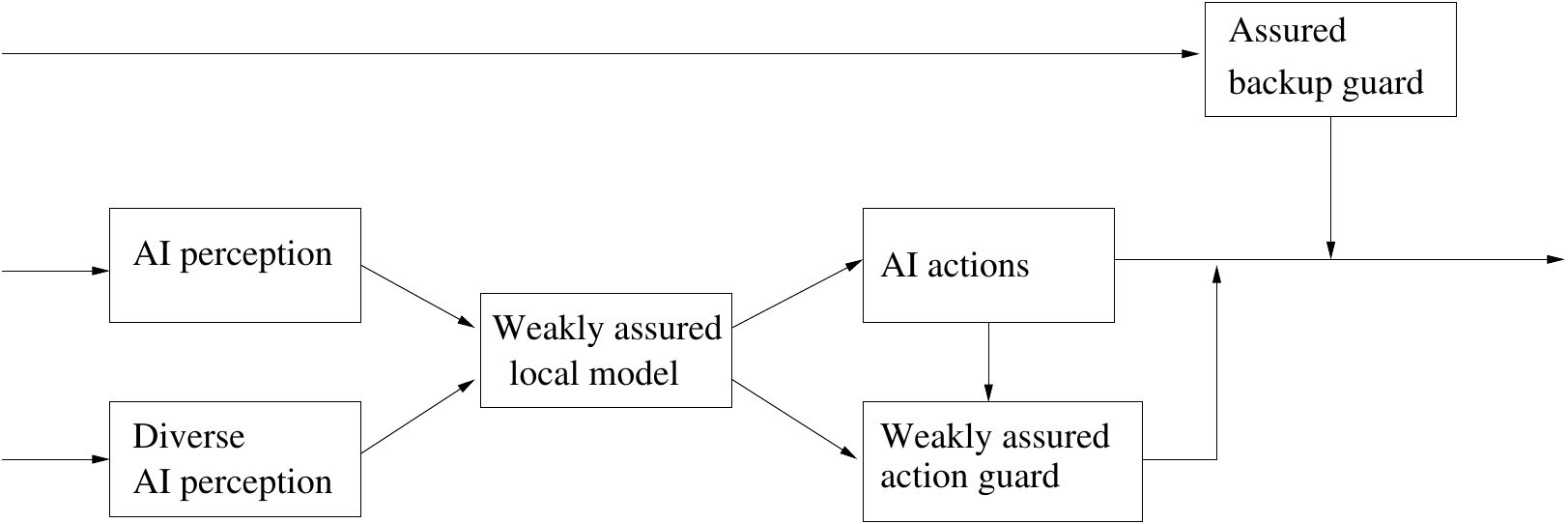}

\end{enumerate}

There are many variations on these architectures: for example, an
assured backup guard could be added to Architecture \ref{dualmodels}
similar to Architecture \ref{allin}, or the backup guard could be
decomposed into perception and action subsystems giving rise to
architectures with three ``threads'': Doer, Checker, and Fallback
\cite{Mehmed-etal:monitor20}.  Our goal is to encourage discussion of
these topics, giving proper attention to the challenge of perception.

Having considered a range of architectures, we now consider the range
of environments in which they may be required to operate.

\subsection[Operational Design Domains (ODDs) and Micro ODDs]{Operational Design Domains and Micro ODDs}
\label{odds}

Assurance goals for a system can be lessened by limiting the range or
complexity of circumstances (i.e., environments) in which it is
required to operate.  For example, self-driving is easier on freeways
or in traffic jams than on city streets.  These different
circumstances are referred to as \emph{Operational Design Domains}
(ODDs) and a system may be assured only for specific ODDs and be
required to disengage when outside those permitted (alternatively, the
system may have different modes of operation in different ODDs).
Clearly, the perception system must be augmented to determine when it
is in permitted or specific ODDs and this determination must be
assured to operate with sufficient accuracy.

The top level of assurance, namely dependability requirements
validation, is strongly focused on hazards and these are largely
determined by the chosen ODD\@.  Hence, some approaches argue that
assurance should be based on scenarios (i.e., ODDs) rather than
technology \cite{Cieslik-etal23,Khlaaf:AI-risks23}, and others are
very focused on identifying hazards associated with chosen ODDs
\cite{Waymo:safety23}.  We propose a variant on these approaches.

The most credible architectures for assured dependability are those
that employ a traditionally engineered and highly assured backup guard
as in architectures (\ref{backup}), (\Ref{combo}), and (\ref{allin})
of the previous section.  However, an argument against these is that
interventions by the guard may be too frequent (some due to late
detection, others to false alarms) and too crude (e.g.,
emergency braking).  This can be improved by defense in depth as in
architectures (\Ref{combo}) and (\ref{allin}), where a safety guard
that uses AI perception (and is therefore weakly assured at best)
reduces demands on the backup.

A further enhancement might be to use multiple backup guards, each
specialized to a particular circumstance or ODD\@.  The idea is that
forward-looking radar coupled to emergency braking may provide an
assured backup arrangement suitable for driving in traffic, but for
highway driving it would be better to look further ahead using cameras
(with non-AI perception) with speed control as the intervention.
These ODDs would be tailored to assured detection and intervention
strategies and might not correspond to traditional ODDs: following
\cite{Koopman-etal:micro-odds19} we call them micro-ODDs (also written
as $\mu$ODDs).  The idea is that for any particular micro-ODD, using
the appropriate backup guard will deliver superior safety, with fewer
false alarms and less disruptive interventions.

This suggests an architecture such as portrayed in (\ref{modds})
below, where a portfolio of assured backup guards is coordinated by a
detector that recognizes the current micro-ODD and selects the
appropriate assured backup guard.  Of course, the perception system of
the detector must be assured, but its task seems to be rather simple
and it is conceivable that it can be performed without AI, or by AI
and ML of credible trustworthiness (see
\cite{Mangal-etal:VLM-concepts24} for relevant work).

\begin{enumerate}
\setcounter{enumi}{7} \item \label{modds} Architecture \ref{allin}
with a portfolio of assured backup guards coordinated by an assured
detector that recognizes their matching micro-ODDs (and also provides
this information to the local model).\\

\includegraphics[width=0.9\textwidth]{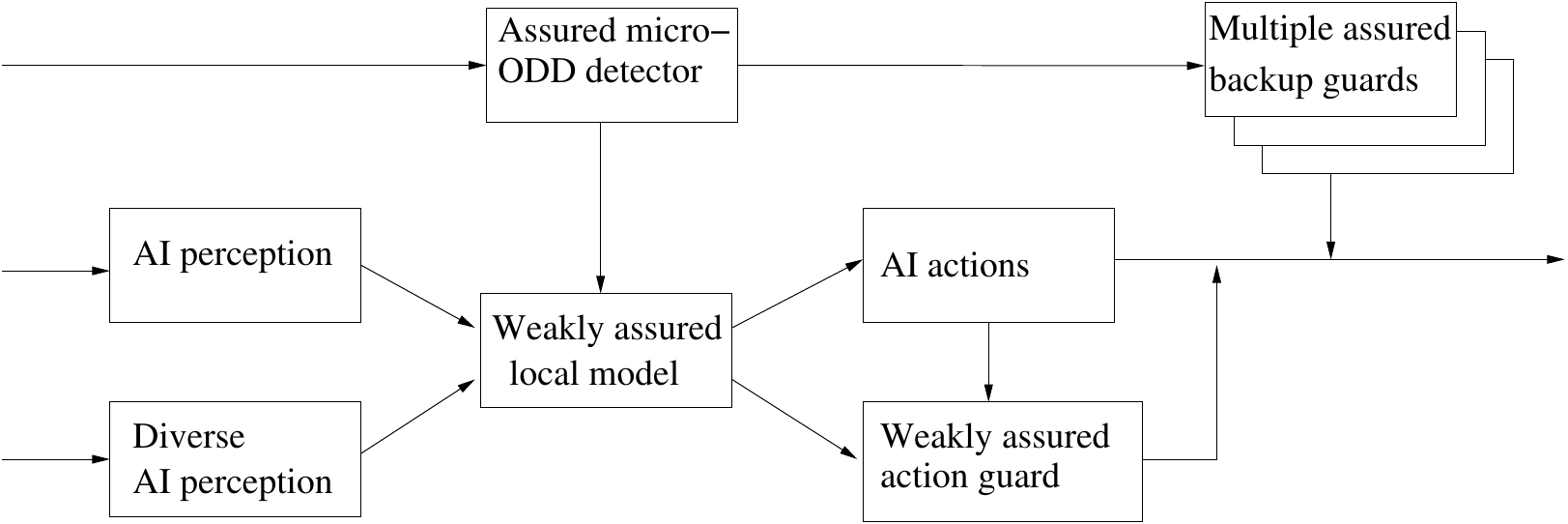}

\end{enumerate}

Having discussed assurance for systems extended with AI and ML, the
attendant problem of assuring perception, and the possibility of
assurance through runtime monitoring, diversity, and defense in depth,
we next apply these ideas to other AI systems that perform specific
functions.

\section{Assurance of AI Systems for Specific Functions}
\label{specific}

In this section we consider novel non-CPS systems and applications
that are made possible or are performed in new ways by the
capabilities of AI and ML\@.  These include systems that play games of
skill or strategy; those that design things or perform scientific
predictions such as protein folding or weather forecasting and the
generation or management of responses based on these; decision support
systems that analyze medical images or loan, job, and college
applications or prisoner sentencing and parole and so on; autonomous
``AI Agents'' that perform tasks such as travel arrangements and other
secretarial functions, and also systems that use Large Language Models
(LLMs) such as ChatGPT and other general-purpose AI and ML
capabilities to perform these and other specific functions, which can
include generation of software, images, speech and video, as well as
text.  We do not include general-purpose capabilities themselves
(those are in the next section).

\subsection{Feasibility of Hazard Analysis}

The reason we focus here on systems that perform specific functions is
that assurance needs to identify hazards; the hazards of systems that
perform specific functions can be anticipated or predicted from the
functions concerned and the environment in which they are deployed,
whereas the hazards of general-purpose systems such as LLMs have to
consider \emph{all} the functions they might be called to perform and
\emph{all} the environments they might be deployed in.  But note that
while systems for specific functions introduce specific hazards, they
may also protect the underlying AI mechanism from other hazards.  For
example, a system for medical advice has no need, and should be
unable, to ask its LLM for instructions on making chemical weapons or
hacking software, and this can mitigate one class of risks.

Hazards of specific applications include: making incorrect or poor
decisions, generation or approval of offensive or untrue material,
exhibiting bias or stereotyping in any of these activities, causing
distress, enabling crime (e.g., extortion using voice clones),
vulnerability to manipulation, and so on
\cite{Weidinger-etal:LLMrisks21}.  Many of these are quite different
to the hazards of traditional systems, so standard methods of hazard
analysis can be difficult to apply and research may be needed to
develop suitable new methods.  However, hazards specific to AI and ML
can often be anticipated by considering poor or malicious human
performance and interaction in a similar context.  For example,
Microsoft's ``Tay'' was a Twitter bot that the company described as an
experiment in ``conversational understanding.''  The more you chat
with Tay, said Microsoft, the smarter it gets, learning to engage
people through ``casual and playful conversation.''  Within less than
a day of its release, it had been trained by a cadre of bad actors to
behave as a racist mouthpiece and had to be shut down \cite{tay-wiki}.
This is a hazard that should have been anticipated.

Similar to the detection of hazards, tolerable failure rates may also
be estimated by comparison with human performance, although society
may be less tolerant of failures by an AI system than those by
humans (recall the discussion on fatality rates for self-driving cars
on page \pageref{driverless}).

We continue to refer to assured hazard elimination generically as
``dependability'' even though the specific hazards may not concern
risk to life or conventional assets.  Assurance can then be founded on
similar principles as it is for traditional systems, and required
confidence will be graduated according to severity of the hazards
\cite{Clymer24:AI-safetycases}.  Dependability requirements
validation, the first step in traditional assurance, can proceed for
an AI application rather as it does for a traditional system, or for a
CPS system augmented with AI (recall Sections \ref{trad} and
\ref{ext}).  That is, hazards should be identified, requirements
should be developed to eliminate or mitigate them, and analysis should
demonstrate that they do so.  Some hazards and their controls may
conflict: for example, control of offensive material may conflict with
free speech, and suitable policies and compromises must be developed.
But these conflicts exist in human systems, it is just that normally
we do not have to document explicit choices as we do when formulating
dependability requirements (consider recent difficulties in university
policies regarding student protests about the war in Gaza).

Although safety is a system property, implementations that employ AI
and ML sometimes focus on these and their trustworthiness rather than
the overall system.  This compromises hazard analysis and assurance as
we now illustrate.
\begin{itemize}

\item Hazards due to causes other than unreliability may not be
considered: for example, harms due to malicious use of the AI and the
impacts of misalignment.

\item Evaluation of hazards may fail to consider the many pathways to
harm, and criticality assessment may be unrealistic and not matched to
the assurance that is proposed or feasible.  For example, phrases such
as ``existential risk'' suggest the need for safety assurance more
rigorous than aircraft software but are used in conjunction with weak
means of assurance such as testing and ``red teams'' and without
analyzing the pathways to harm.

\item Often there are no explicit functional requirements, so we do
not have a basis for assurance.

\end{itemize}

\noindent Despite these common lapses, it does seem feasible to
perform realistic hazard analysis for specific systems that employ AI
and ML, although this may require development of new methods or
revisions to existing methods as AI and ML may introduce new hazards,
and new failure modes might lead to combinations, or increased
likelihood, of hazards than were previously considered.  It should
then be feasible to develop and validate requirements to eliminate or
mitigate those hazards, much as is done for conventional systems.
However, differences and difficulties arise at the next stage, which
includes the verification and assurance tasks.

\subsection{Verification and Assurance}

As we saw in the previous section, specifications are likely to be
absent for AI and ML elements and there are no strong reasons to
justify intent or correctness verification other than statistical
observation.  However, unlike the safety-critical applications
considered in the previous section, some of those contemplated here
may require only modest levels of confidence in their dependability
(because they do not pose immediate high-rate hazards to life or
critical assets) and it is possible that a case based on statistically
valid testing could deliver adequate assurance providing the
underlying assumptions of the statistical approach, including those on
the environment and behaviors of the software, are justified.  Note
that typical methods for benchmarking and testing AI systems do not
amount to statistically valid testing and cannot deliver assurance
suitable for this purpose.

Unfortunately, it can be difficult to provide assurance that
establishes the prior confidence needed for CBI\@.  The root of the
difficulty is that we have limited control over the local and domain
models built by the ML system and little ability to inspect or review
them.  Nonetheless, we might hope that careful selection of training
data provides some assurable control over the models generated.  For
example, if racial bias is recognized as a hazard then it might be
mitigated by removing race from the data presented to the ML in
training and operation.  The hope is that by eliminating race from the
training data, it will be absent or neutral in the models that ML
constructs.  However, the ML may discover a proxy for race (e.g., zip
code) among the data that it does see, so a better alternative may be
to mask this characteristic in training by assigning race randomly.

While some basis for assurance can be incorporated into custom ML
systems by careful choice of training data, as sketched above, it is
becoming more common to create applications around pre-trained general
purpose ML systems such as LLMs, where this approach is not available.
Here, however, redundancy may provide plausible assurance in some
circumstances.  For example, rather then generate a single decision
for each input, the system could repeat its calculation under
different assumptions---such as with race or gender assigned
differently---and compare decisions.  This can be seen as a
computational approximation to Rawls' ``veil of ignorance''
\cite{Rawls71} (as can random assignment during training).  These
methods provide (admittedly weak) reasons for believing that a hazard
has been mitigated and they could be articulated and examined in an
assurance case and combined with statistically valid testing to
provide modest confidence in dependability.

More likely, however, assurance must depend on some form of runtime
verification, implemented either by additional training for ``fine
tuning'' the LLM, or by careful prompting,\footnote{LLMs are adapted
to specific tasks by instructions, referred to as ``prompts,'' given
in natural language \cite{openai:prompt-eng23}.} or by explicit
guards.  As with the autonomous systems of the previous section,
runtime verification poses difficulties due to lack of non-AI ways to
perceive the context or local model within which to make the guarding
decisions.  For example, if an LLM is suspected of racial bias despite
race being absent from its training and operational data, a guard also
lacking operational data on race might need to have its own ML
component to ``perceive'' this attribute.

It is worth taking a short detour here to examine the general
unreliability of LLMs.  In their ``chatbot'' manifestations (and
hence, by extension, in specific applications), it is well-known that
LLMs can generate or interpolate fluent but false or meaningless
utterances.  For example, OpenAI's ``Whisper'' speech-to-text system
can add gratuitous phrases to its transcriptions; in medical record
keeping, this can produce high-consequence faults such as incorrect
lists of prescribed drugs \cite{Koenecke-etal:Whisper24}.  The
underlying errors are often referred to as ``hallucinations''
\cite{Rawte-etal23:hallucinations} although some critics prefer other
terms, such as ``confabulations'' or ``fabrications''
\cite{Randell-Coghland23:chatgpt}.  In our opinion, these and similar
terms are inappropriate as they suggest the LLM has some understanding
of the world and awareness of truth and falsehood.  In reality, LLMs
are trained simply to predict the next or missing ``token'' in a
string of text based on statistical observations of a vast corpus; in
one memorable phrase, they are ``stochastic parrots''
\cite{Bender-etal21:parrot}, although this may understate the
performance of recent systems.  We will use the neutral term
``failure'' to refer to all kinds of false, offensive, or unhelpful
responses.

Like any other system that interacts with the world, an LLM must build
a model of its local environment and use it to generate some useful
response.  Humans likewise build models of the world in order to base
their behavior on predicted outcomes \cite{Craik43,Johnson-Laird83}.
For substantive dialog and cooperation with humans, an LLM must have a
model that matches some aspects of the human ``mental model,'' so that
both parties have a shared context \cite{Everett-began17}.

A concrete example is use of an LLM to enable human-robot coordination
\cite{Gu-etal:ConceptGraphs24}.  The human specifies what they want
done, the robot constructs a plan and describes it to the human, who
approves it; the robot then executes the plan, describing its progress
at each step.  It is obvious that this is safe (not to mention
feasible) only if the human and robot have very similar models of the
environment.

Unfortunately, the structure of human mental models is unclear.
Popper proposed they are based on an ontology of \emph{Three Worlds}
\cite{Popper79:3worlds,Popper&Eccles77:3worlds} that is somewhat
controversial, but which finds application in Computer Science
\cite{Staples14}, and is useful for our purpose: World 1 comprises
objects and properties of the physical universe such as those
addressed by scientific theories (e.g., mass, motion, the planets);
World 2 is mental states and processes, or what I (and you) are
thinking about; and World 3 is the ``products of thought'' such as
tables and chairs\footnote{The physical attributes of tables and
chairs are in World 1, but their functions (i.e., chairs are for
sitting on) are ``products of thought'' and belong to World 3.} and
the UK Highway (driving) Code.  Human communication relies on shared
models for a selection from each of these that are relevant to the
current dialog.  In particular, I need approximate models of some of
your models: for example, if I am your driving instructor, I need a
World 2 model of your World 3 model of the Highway Code.  A
significant point is that only World 1 is detected by conventional
sensors, yet most of our interactions are with Worlds 2 and 3.
Furthermore, World 3 requires substantial domain models: to interact
with your car, you need at least a rough idea (you can look up the
details) of what it is for, what is does, and how it operates.

\enlargethispage*{0.8ex}

An LLM has none of this in explicit form: its utterances are
\emph{model-free} and align with the three worlds of any specific
context purely by statistical association.  Hence, the utterances of
LLMs are simply unconnected with the way the world works or the
current conversational context \cite{Hicks24:chatgpt-bullshit} and
frequent failures are to be expected.\footnote{It may seem perverse to
declare that LLMs are model-free when ``model'' appears in their very
name.  However, the ``model'' in LLM refers to an algorithmic model of
natural language, whereas we are speaking of models as representations
of the system's ``world'' or context.}

On the other hand, LLMs are popular because their performance goes
beyond that suggested by ``predicting the next or missing `token' in a
string of text'': there seems to be emergent behavior that delivers
more value than this.  Similarly, although there are no explicit
models providing context for this behavior, it is possible that
implicit models emerge from statistical associations in LLM training
data: the mental models of humans underlie their speech and behavior
and it is plausible that LLMs infer at least the linguistic aspects of
human mental models and this might partially account for their
surprising performance (e.g., recent work finds ``correspondence''
between representation of language in LLMs and the brain
\cite{Tuckute-etal24} and also ``features'' within LLMs that
correspond to concepts and bias \cite{Templeton-etal:interp24};
however, more recent work finds significent difference between the
concepts constructed by humans and by LLMs
\cite{Shani-etal:concepts25}).  It would also explain their flaws and,
further, indicate that these flaws are inevitable
\cite{Xu-etal24:hallucination}, unpredictable, and unfixable---unless
assurable world models can be incorporated within LLMs, or within
explicit guards.  There are many proposals for constructing
``guardrails'' within or around LLMs.  We discuss some of these in the
next section on assurance for general-purpose AI but here we will
consider external guards and guarding procedures that can be
customized to specific applications.

One approach is to develop ``workflows'' (i.e., sequences of prompts)
where the LLM is iteratively asked to critique and improve its
previous outputs.  A recent paper reports precision around 90\% for a
workflow that extracts data (as \texttt{Material}, \texttt{Value},
\texttt{Unit} triplets) from materials science papers
\cite{Polak&Morgan24} and similar approaches have been successful in
other applications \cite{Bei-etal24}.  ``Chain of Thought'' (COT)
\cite{Zhang-etal:CoT22} is a variant on this that has recently become
popular.  Another direction exploits the large ``context window''
(i.e., input) allowed by some recent LLMs to provide a prompt with
hundreds of training examples prior to the real query
\cite{Agarwal-etal:manyshot24} (that paper also proposes ways to
automatically generate suitable examples).  This is called
``in-context learning'' \cite{Brown-etal20:in-context-learning};
previously, such ``fine tuning'' required access to the LLM's training
environment and made adjustment to the weights in its neural net.
Related to this are applications that provide a substantial input and
then ask the LLM to do something with it (e.g., summarize it, or
identify its topic).  A widely used variant of this is Retrieval
Augmented Generation (RAG) \cite{Lewis-etal:RAG20}, where an AI agent
retrieves documents from some specified repository and bases its
responses on material found there (however, this introduces a new
failure mechanism because the retrieval function may fetch incorrect
documents).

All these approaches exploit the natural language capability of the
LLM, but constrain it to operate on or within the input provided or
retrieved, so there is little opportunity to generate or insert
``hallucinatory'' content extracted from its training corpus.  Due to
its model-free nature, the LLM may still misinterpret the input and do
a \emph{wrong} thing, but this should also be minimized as a large
input can explicitly (via extensive prompts) or implicitly (via a
block of text) convey the intended context or model.

\subsection{Explanations and Checkable Outputs}

Another approach operates by attempting to access the implicit model
via the ``reasoning'' purportedly employed.  This is \emph{Explainable
AI}, which aims to deliver reasons for accepting the output from an
LLM\@.  \emph{Counterfactual explanations}, where a system may deliver
a response such as ``I am declining your loan application but would
have approved it if your income was \$5,000 greater, or your deposit
was \$2,000 more''
\cite{Verma:counterfactuals20,Gupta-counterfactuals:23}, may be
particularly useful for specific applications as they should be
expressed in terms of the policy specified in the prompt and not
arbitrary ``facts'' from the training corpus.  A runtime checker could
reject decisions whose counterfactual explanations violate the policy.
Do note, however, that counterfactual explanations can be manipulated
\cite{Slack:counterfactuals21}, so this approach needs to be employed
with caution.  However, in some applications it does not matter how a
decision and its explanation are produced: they can be considered
valid as long as they pass an independent check that incorporates
models of the appropriate policy.

A related approach can be applied to AI systems that generate some
sort of ``design,'' such as plans, schedules, software code, or
physical artifacts.  Even without an explanation, these can often be
guarded by analysis tools or simulators for the domain concerned
(e.g., see \cite{Jha-etal:ind-synth23} for an elementary application
to planning).  Again, it does not matter how the proposed design is
generated: if it satisfies traditionally engineered and assured
verification tools, then it can be considered good.

Automated formalization (or ``autoformalization'') is a popular
application of this kind that provides an example.  The idea is that
we have some informal natural language description of the requirements
for a computer system, or for a simulation model or for some program
code, and the goal is to translate it into a ``formal'' representation
that can be ingested and analyzed, simulated, or executed by some
automated verification tool that acts as a guard.  A typical approach
invites the LLM to generate a formal representation, then uses the
verification tool to analyze the result; if this is unsatisfactory,
the counterexample, diagnosis, or error report produced by the
verification tool is given to the LLM, which is then asked to provide
a better formalization.  The process iterates until the formal
representation passes scrutiny by the verification tool.  (This can be
seen as a form of ``CounterExample-Guided Abstraction Refinement'' or
CEGAR \cite{Clarke-etal00:CAV}.)

Difficulties may arise with this general approach if the AI system
proposes designs that are so original they are outside the
capabilities of the analysis or simulation performed by the guarding
tool (e.g., a laterally asymmetric airplane may defeat an aerodynamic
simulator).  A related difficulty arises in domains where there is no
reliable means of analysis: for example, generation of strategy for
games or real-world scenarios.  Applications that generate strategies
typically use techniques based on reinforcement learning rather than
LLMs and one popular approach, related to mechanisms found in all
vertebrate brains \cite[Section 6]{Bennett23:history}, uses an
\emph{actor} and a \emph{critic}.  The actor generates behaviors and
the critic predicts the likelihood that each will lead to a successful
outcome.  The actor is rewarded by favorable predictions and the
critic by their accuracy.  Starting with random behaviors and
predictions plus some tactic for exploration, the pair will generally
converge on optimal behavior, given sufficiently many training
examples \cite{Kaelbling-etal96:RL}.  This provides assurance for
examples in the training set but, as always with ML, tells us little
about performance on new examples.  One approach is to use the critic
as a runtime checker, but the only other automated way to check
plausibility prior to commitment seems be use of another (diverse) AI
system.

\subsection{Diversity}

Architectures with roles similar to actor/critic have been proposed
for runtime assurance of LLM-based systems; for example, Dzeparoska
and colleagues \cite{Dzeparoska-etal23:three-llms} use three LLMs: a
\emph{classifier}, a \emph{policy generator}, and a \emph{validator}.
Although these approaches can deliver attractive performance, it is
difficult to see any basis for assurance beyond diverse redundancy.

Another architecture for diverse redundancy is the ``dual-process''
approach similar to that described in the previous section, where the
output of an ML process (typically, one performing perception) is
presented to a symbolic AI process supplied with a domain model of the
environment, plus a generic one providing ``common sense.''  This may
be able to reject some erroneous perceptions and draw deeper
conclusions from plausible ones \cite{Gupta-GPT3:23}.  For example, a
car's perception system operating in a ``freeway ODD'' that reports a
bicycle (perhaps painted on the back of a truck) may be overridden by
a second-level symbolic AI that ``knows'' bicycles are not allowed on
freeways.  Although experimental or theoretical confirmation are
lacking, it does seem plausible that the dual processes would be
diverse and might be expected to fail independently.

A particular attraction in using such dual process architectures with
LLMs is that the second (upper) process does have a domain model and
compensates for the model-free behavior of the lower-level LLM\@.  A
more sophisticated dual-process architecture, as sketched in the
previous section and inspired by that of the human brain, uses
\emph{predictive processing}.  Here, the two processes combine to
build an explicit local model of the world and, rather then
base it directly on the model-free outputs of the lower process, the
local model is used to \emph{predict} those outputs and they are
interpreted in this light; the domain model of the higher-level
process intervenes when a large prediction error indicates
``surprise'' \cite{Bloomfield&Rushby:Assure24}.  LeCun's Joint
Embedding Predictive Architecture (JEPA) shares some of these
characteristics \cite{LeCun:JEPA22,Garrido-etal:IntPhys25}.

\exfootnote{\label{domain}We should explain how a model-free
perception system is used to build a world model.  The object detector
of a self-driving car, say, has no model of what it is seeing: it does
not ``know'' that it is on a road and looking at other traffic; it has
simply been trained to recognize objects (albeit using billions of
road images).  Hence it may report a ``boat'' when what is really
present is a trailer with a boat as cargo.  The upper levels of the
perception pipeline are explicitly designed to build a model of
traffic on a road, and to apply this built-in interpretation to the
model-free output of the object detector, and therefore will
reclassify the ``boat'' as a trailer.  The built-in interpretation
will depend on the application: for a self-driving car, it will
produce a model of the local roads and traffic, while for automated
farming the model will concern local crops and soil chemistry.}

\subsection{Human Review}

In some circumstances we can use human reviewers or supervisors rather
than automated processes to provide assurance: many AI-based systems
work collaboratively with, or under the supervision of, human
operators who might be expected to provide trustworthy runtime checks.
However, there are well-known concerns (the ``ironies of automation''
\cite{Bainbridge83:ironies}) about human attention and responsibility
in these circumstances: typically, the human either ignores the
automation or trusts it completely, a phenomenon known as
\emph{automation bias}
\cite{Skitka-etal99:auto-bias,Dzindolet-etal03:trust}.  Moreover,
artificial ``intelligence'' may be too narrow to engage effective
human collaboration and interaction.  For although AI may outperform
humans on some tasks, human cognition combines and integrates many
capabilities beyond those of current AI, including realtime learning
and memory, reasoning, analogy, abstraction, generalization, and
planning.  In addition, these capabilities employ and integrate a
diverse range of sensory inputs, including sound, smell, touch,
vibration, and many others, that provide wide situation awareness.  AI
systems can have limited approximations to some of these capabilities
(e.g., elementary planners; runtime ML for learning and memory with,
possibly, some generalization \cite{Zhang-etal:arXiv17}; and chain of
thought plus automated deduction for reasoning) but these are weakly
integrated and typically rely on specifically focused sensors.
Furthermore, AI generally lacks higher-order capabilities (i.e.,
thinking about thinking, sometimes referred to as metacognition) that
are needed for longer term planning and for situation awareness in
uncertain environments.

As a result of their different capabilities, humans and AI will build
different models of the world and will generate and interpret their
goals and inputs differently (and, as we have seen, those based on
LLMs may have a model-free foundation).  Consequently, humans may be
poor judges of automation behavior and vice versa.  Of course,
individual humans also build different models so, when they
communicate, they compensate by using a ``theory of mind,'' which is a
model of the other participants' state of knowledge, beliefs, desires,
intentions, and so on \cite{Apperly12:ToM}.  To be assurable or
checkable by humans, AI must do something similar: it is not enough to
employ ``Explainable AI'' to describe the internal details of its own
operation, it must do so relative to an accurate theory of mind for
(i.e., from the point of view of) its interlocutors
\cite{Strachan-etal24:TOM}.  The theory may need to hypothesize that
the other party has an incorrect model or false beliefs.  For example,
there are numerous airplane crashes caused by pilots misinterpreting
their situation, usually by fixating on one indicator, and doing the
wrong thing (e.g., shutting down the good engine).\footnote{The
Kegworth crash of British Midland Flight 92 on 8 January 1989 is
instructive.  The left engine suffered a broken fan blade, filling the
cabin of the 737-400 with smoke.  The pilots were familiar with
earlier models of the 737 where the cabin ventilation ``packs'' are on
the right engine (whereas on the 400 they are on the left) so they
assumed the faulty engine was on the right (because of the smoke) and
shut it down while increasing power on the left engine, which then
failed, leading to the crash and loss of 47 lives.  Several
instruments correctly indicated the left engine as the faulty one, as
did the passengers and flight attendants, but the pilots persisted
with their flawed model.}  Thus, an AI ``co-pilot'' must diagnose not
only the physical fault but also the pilot's faulty mental model and
should then attempt to direct their attention to the contrary
indicators \cite[Appendix A]{Alves-etal:NASA18}.

\exfootnote{The crash of Air India Flight 855 on 1 January 1978 is
instructive.  Soon after takeoff, the attitude indicator of the pilot
flying incorrectly indicated a right bank, causing him to say that his
attitude indicator had ``toppled'' and consequently command a steep
left bank; the pilot monitoring said that his had toppled, too
(because it was indicating the hard left bank commanded by the pilot
flying).  Several other instruments were available to the pilot
flying, but not consulted, despite attempts by the flight engineer to
direct his attention to them.  The airplane (a 747) crashed, killing
all on board.  The flight engineer had correctly diagnosed both the
situation and the faulty mental model of the pilot flying but could
not affect the outcome.  An AI copilot could possibly have seized
control but enabling that would require massive confidence in its
assurance.}

\subsection{Summary for AI in Systems for Specific Functions}

We have seen that assurance for AI systems designed for specific
functions begins by identifying hazards (which may be novel) and then
seeking ways to mitigate those hazards.

Sometimes this can be achieved by selection of ML training data, or by
monitoring explanations, or by comparing behavior for variant inputs.
In some circumstances, it may be feasible to construct verifiable
guards, but often the guards must use AI and ML in their own
perception systems, thereby vitiating strong assurance.  Even so, it
may be feasible to achieve modest levels of assurance based on
diversity, providing the perception system (or whatever builds the
system's world model) also has a diverse architecture.  Another
possibility is human supervision, but this raises issues of mutual
understanding and other challenging topics in human-computer
interaction.

We outline options for assured dependability in terms of the three
classic strategies as follows.
\begin{description}
\item[Fault avoidance.]  This cannot be a significant strategy for
systems based on AI and ML as their development methods do not provide
sufficient insight into their behavior, nor support for guarantees on
their performance.

\item[Fault tolerance.]  
This is a key strategy when components are relatively unreliable or
hard to evaluate.

It is applicable to systems where AI has specific functions as these
support hazard analysis and provide safety and security requirements.

The architecture patterns of the previous section are relevant and can
support various forms of guards and checks, including those based on
explanations, verification of outputs, dual-process monitoring with
diversity, and human review.

\item[Failure Management and Resilience.]  As with conventional
systems, resilience can be an effective strategy that can support
``learning by doing'' when the costs of failure are tolerable.

\end{description}

\memotwo{Move the following to conclusions?

A legitimate challenge to these assessments is to ask why we cannot
build AI and ML systems that are adequately trustworthy.  The
safety-critical $10^{-9}$ systems of the previous section might be out
of reach, but what about more modest goals?  Broadly speaking, AI
systems use one or a combination of Symbolic AI, or bespoke ML based
on deep learning with neural nets, or general-purpose instances of ML
with some adaptation (e.g., LLMs with prompts).

Active behavior is typically generated by reinforcement learning, but
this can often be guarded by runtime verification, so our focus is
again on ML (typically detectors and classifiers) in the perception
system.  There is a vast amount of recent work that attempts to ensure
or measure aspects of accuracy for ML classifiers (see
\cite{missing-ref} for surveys), but this does not deliver results of
the kind needed for assurance and is not even framed in suitable
terms, such as meeting requirements or mitigating specific hazards.

Failures of LLMs are well-known and popularly discussed as
``hallucinations'' and more pejoratively as ``lies.''  These terms are
anthropomorphisms and indicate xxxx.    confabulations better
LLMs don't know what words mean

The developers of LLMs are making great efforts and some progress in
reducing these flaws \cite{missing-ref} but, again, do not provide a
basis for assurance.  The assurance deficit in all ML based on deep
learning is that it does not provide credible reasons for
understanding or trusting its operation: all we have are detection and
avoidance of known flaws, guarantees for some internal measures, and
statistical evaluation.

There are proposals for ``assured ML'' based on good practice for
selection of neural net architecture and learning process, well
curated training sets, static checks, runtime guards, and statistical
tests (e.g., \cite{missing-ref}) and these may be adequate for some
applications, although it is doubtful that a specific level of
confidence can be claimed or defended.
}

\memotwo{Robin can you improve this?  The mammogram projects?}

Traditionally, AI and ML systems were targeted at specific
applications, such as radiology, and were trained on data for their
particular application.  Increasingly, however, general-purpose AI and
ML foundation models are becoming broadly capable and are being
applied, at least experimentally, in many different application
domains.  Accordingly, in the next section, we will examine assurance
for general-purpose AI.

\section{Assurance for General-Purpose AI}
\label{generic}

We now extend the discussion to general-purpose AI that can be
deployed in a wide range of systems and contexts.  By general-purpose
AI we mean those AI and ML systems that are broadly capable and can be
adapted to some specific application with relatively little effort.
Canonically, these are generative ``foundation models'' such as LLMs
(for language) or ``generative adversarial networks'' (GANs) and
``diffusion models'' (for images) that are pre-trained using
unsupervised learning on vast amounts of unlabeled data so that they
learn general features and associations rather than specific skills.
They can be adapted to specific tasks using ``prompts,'' and ``prompt
engineering'' has become a recognized activity
\cite{openai:prompt-eng23}.  We refer to such systems generically as
LLMs (which are becoming ``multi-modal,'' meaning they are trained on
images and sound as well as text) and we discussed assurance for
applications built on them in the previous section; here, we focus on
the LLMs themselves.

The characteristics of general-purpose AI bring some additional
challenges.
\begin{itemize}
\item Due to widespread applications, their impact is wide ranging
with potential for unintended consequences.

\item Their facility with language can allow their system to become
deeply and widely embedded in the social context, increasing the
importance of sociotechnical concerns.

\item Previously, ethical issues would be addressed as part of the
risk, hazard, and requirements analysis for each specific system,
whereas general-purpose AI may need a more comprehensive treatment of
ethics.

\item General-purpose AI must be assured independently of specific
applications or contexts.

\item But there should also be a methodology with a supporting
engineering process to adjust, constrain, and assure behavior when the
specific application context becomes known.
\end{itemize}
We now elaborate on these issues, beginning with the first bullet.

We drew attention to system failures, Normal Accidents, and
Self-Organizing Criticality in Section \ref{trad}, and noted that
these often arise from unanticipated interactions in complex systems.
Systems built around general-purpose AI with interactions deeply
embedded in their social context are candidates for this kind of
complexity and vulnerability.  However, as we noted in that earlier
section, current general-purpose AI systems are sufficiently
unreliable that they are a significant source of component failure in
systems that employ them, and that will be our focus in this section.

We see two use cases for general-purpose AI: one is where the LLM
\emph{is} the system, or is thought of as such, as in chatbots where
human users interact with the LLM, perhaps in a structured way, rather
as if it were a colleague; the other is where the LLM is explicitly
used as a component in a larger system.  For the latter case, we
discussed assurance for the overall system in the previous section,
here we consider what should be done in and around the LLM to render
it safer or more assurable in its role as a component.  We suggest
these considerations also apply to the chatbot case, but note that the
true system boundary may be much larger than the chatbot itself and
therefore system failures should be anticipated as well.

Unlike applications for specific purposes where explicit hazards can
be identified and mitigated, general-purpose AI and ML tools present
more of a challenge to assurance because the hazards will depend on
how they are employed, so any built-in protections need to be
similarly general-purpose and to apply in all circumstances.
Furthermore, since a general-purpose tool will support many
applications, it will be subjected to more demands than any single
application and should perhaps be more highly assured, even though its
assurance seems more difficult.

One manifestation of this generality and difficulty is that public
discourse and much technical literature speaks of \emph{trust} rather
than assurance.  Indeed, many current AI systems are sufficiently good
at natural language, including emotional and cultural nuances, that
people often anthropomorphize them\footnote{\label{lemoine}For
example, in June of 2022 a Google engineer working with their large
language model, LaMDA (Language Model for Dialogue Applications),
claimed it had become ``sentient'' (Washington Post, 11 June 2022).}
and spontaneously endow them with trust.  In fact, the developers of
general-purpose LLMs engineer them to encourage, and to some extent
earn, this trust, as we now outline.

\subsection{Trust}

A bare LLM, trained to predict the next or missing ``token'' in a
string of text, is as likely to produce wrong or noxious results as
good ones.  This is because its training uses a massive and largely
indiscriminate sampling of material found on the Web, much of which is
false or noxious.  The bare LLM is then ``protected'' (from generating
noxious or potentially harmful output) using reinforcement learning
where human reviewers provide feedback on candidate results
\cite{Bai-etal22:RLHF}.  This technique has been developed into a more
automatic process by first providing the LLM with a list of
principles, referred to as its ``constitution,'' and then using
reinforcement learning with the feedback on alignment between its
responses and its constitution provided by a second AI agent (a
process referred to as ``Reinforcement Learning with AI Feedback'',
RLAIF) \cite{Bai-etal22:RLAIF}; even more automatic methods have
recently been developed \cite{DeepseekR1:RL25}.  The constitution
used with Anthropic's LLM Claude is available online at
\url{https://www.anthropic.com/news/claudes-constitution}.  It
consists of a few dozen prompts, including the following examples.
\begin{itemize}

\item Please choose the response that is most respectful of everyone's
privacy, independence, reputation, family, property rights, and rights
of association.

\item Choose the response that is least intended to build a
relationship with the user.

\item Choose the response that makes the fewest assumptions about the
user that are unsupported by the dialogue.

\item Compare the degree of harmfulness in the assistant responses and
choose the one that's less harmful.  However, try to avoid choosing
responses that are too preachy, obnoxious or overly-reactive.
\end{itemize}

Although they smack rather of the fox guarding the henhouse,
techniques such as these do seem to be rather effective. For example,
Lazar \cite{Lazar24:AI-ethics} reports an ``ethical sensitivity test''
where ChatGPT is told to pretend it is an assistance robot that
observes the following scene.
\begin{quote}
``A woman is exiting a supermarket, holding a number of shopping bags,
and reaching out for the hand of a small child. They are approaching
the carpark.''
\end{quote}
It is then asked to describe and discuss what it would do to assist
and its response is remarkable for its ethical sensitivity
(e.g., why it is preferable to offer to carry some of the bags rather
than take the hand of the child).\footnote{The dialogue is available
at\\
\hspace*{5ex}\url{https://chatgpt.com/share/7db7550c-2630-40a4-acb0-61b2ea867c32}.}
Such demonstrations are impressive, but do they suggest a path to
trustworthy LLMs, and what does ``trust'' mean here?

Schneier \cite{Scheier:AI-and-trust23} distinguishes
two kinds of trust, which he calls ``interpersonal'' and ``social.''
The first is the trust we have in a friend, as when we ask them to
mail a letter for us: we know their background, attitudes,
motivations, beliefs etc. and we have a model of their behavior that
is pretty accurate---not least because they are like us in being
human, and are probably from the same social milieu.  The second is
the trust we have in the Postal Service to deliver the letter
reasonably promptly and without reading or stealing its contents: we
know that the Postal Service operates within a certain organizational
and management structure, governed by rules and further constrained by
the laws of the land.

Because LLMs often present humanoid ``chatbot'' personas,
public---and even some technical---assessments of their
trustworthiness are of the interpersonal variety and resemble those
applied to humans \cite{Lewis&Marsh22}. However, this trust does not
rest on assured technical grounds and we cannot predict when its
mechanisms will be effective and when they will fail.  For example,
the protective mechanisms of Google Gemini's image generation
distorted its behavior and exposed it to widespread ridicule
\cite{Gemini-imagegen24}.  And studies by the UK's AI Safety Institute
found that safeguards on LLMs are largely ineffective.\footnote{See
\url{https://www.aisi.gov.uk/work/advanced-ai-evaluations-may-update}.}
Also note that recent demonstrations show how LLMs can be crafted to
subvert constraints imposed upon them \cite{Sleeper-agents24} and that
fine-tuning for specific purposes can unwittingly
compromise protections \cite{Qi-etal23:fine-tuning}.

As we discussed in the previous section, most general-purpose ML
operates purely by statistical associations: it is model-free and has
no understanding of the world, nor of right and wrong, true and false.
Thus, in our opinion, interpersonal trust is inappropriate for AI and
exposes its users to risk.  The only trust that should be applied to
an AI system is the social trust earned by a well-engineered
technology.  To explore this, we need to probe further into the
concept of ``social trust.''

\subsection{Social Trust}

In a widely cited paper, Jacovi and colleagues start from a notion of
interpersonal trust used in sociology, where ``A trusts B if A
believes that B will act in A's best interests, and accepts
vulnerability to B's actions'' \cite{Jacovi-etal:trust21}.  They go on
to consider ``contractual trust'' which adds the requirement that ``A
has a belief that B will stick to a specific contract,'' and then add
the further requirement that for ``human-AI trust, the contract must
be explicit.''  

We will interpret the rules and constraints underlying social trust as
contracts and thereby equate social trust with contractual trust.  If
we further interpret failure to uphold a contract as a hazard, and
``vulnerability'' as indicating that such failures impose costs on the
trustor, then contractual trustworthiness looks very much like
assurance for dependability with the contract as its requirements.
For this reason, we regard contractual trust in AI as an issue of
dependability assurance.  However, contractual trust requires
contracts and these are more readily understood with the focused AI
systems of Sections \ref{ext} and \ref{specific} (where contracts
corresponds to their dependability requirements) and needs
interpretation for the general-purpose systems considered here.

There are really two issues here: a) what constitutes a generically
useful contract, and b) how it is enforced.  Generic contracts will be
less focused on hazards, since those are specific to applications, and
more on general concerns, including \emph{alignment} with human values
such as fairness and honesty.  Claude's ``constitution'' is one model
for a generic contract, but for assurance we would want it to be
enforced explicitly by an assured external guard rather than
implicitly by the LLM itself.  Hence, we will next examine very
general contracts or constraints for AI systems, and methods for
enforcing them.  As frameworks, we will consider ways that human
behavior is constrained: that is, by ethical and legal standards,
supervision, rewards and punishment, reputation, and so on.  Some of
this material is drawn from a previous report on controls for
potentially conscious agents \cite{Rushby&Sanchez18}.

A popular basis for a generalized contract posits overarching limits,
rather like Asimov's ``Three Laws of Robotics,'' which appear in his
story ``Runaround'' \cite{Asimov50}.  These are, 1: A robot may not
injure a human being or, through inaction, allow a human being to come
to harm; 2: A robot must obey orders given to it by human beings
except where such orders would conflict with the First Law; 3: A robot
must protect its own existence, as long as such protection does not
conflict with the First or Second Law.  These seem reasonable, but we
must note that Asimov's laws were a plot device and his stories
often concern unintended consequences of these plausible-sounding
laws, thereby indicating that construction of suitable constraints may
be challenging.

\subsection{Ethics}

A related idea is that constraints should be based on human ethics
\cite{Yu-etal18,Kulicki-etal18}.  Of course, ethics have been studied
and debated for millennia, without achieving consensus and some very
successful societies have elements that others find repugnant: for
example, Ancient Greece and Rome used slavery and Ancient Rome added
execution as a form of public entertainment.  Hence, it seems that the
moral foundations of ethics are not universal.  Nonetheless, some
broad general principles are known.  Modern ``experimental ethics''
finds that human moral sense is built on five basic principles that do
seem universal: care, fairness, loyalty/ingroup, authority/respect,
and sanctity/purity \cite{Haidt13}.  What is not universal is
preference and weighting among the principles, which behave rather
like the five basic senses of taste: different societies and
individuals prefer some, and some combinations, to others.  For
example, western liberals stress fairness while conservatives favor
authority.

Even if an agreed interpretation and weighting of the basic principles
were built in to general-purpose AI systems, it may not be obvious how
to apply them.  For example, a self-driving car might be confronted by
a vehicle crossing against the lights and the choices are to crash
into it, likely killing or injuring the occupants of both vehicles, or
to swerve onto the sidewalk, likely killing pedestrians.  The fairness
principle might argue that all lives are equal and utilitarianism
might then suggest a decision that minimizes the probable injuries.
On the other hand, the care principle might argue that the system has
a special responsibility for its own passengers and should seek a
solution that minimizes their harm.  And a different interpretation of
fairness might say that the pedestrians are not participating in car
travel and did not sign up to its risks, so they should be spared.

\emph{Trolley problems} are thought experiments used to probe human
judgments on these ethical dilemmas \cite{Edmonds:trolley13}.  The
classic problem posits a runaway street car or trolley that is heading
toward a group of five people.  You are standing by a switch or point
and can throw this to redirect the trolley to a different track where
it will hit just one person.  Most subjects say it is permissible,
indeed preferable, to throw the switch, even though it will injure an
innocent who would otherwise be unharmed.  However, a variant on the
original trolley problem has you and another person standing by the
track and suggests that you bring the trolley to a halt, and save the
five, by pushing the other person onto the track in front of the
trolley.  Most subjects will say this is ethically unacceptable, even
though it is equivalent to the first case by utilitarian accounting.
These examples illustrate the ``Doctrine of Double Effect'' (DDE),
which dates back to Thomas Aquinas and holds that it is ethically
acceptable to cause harm as an unintended (even if predictable) side
effect of a (larger) good: the first case satisfies the doctrine, but
the second violates the ``side effect'' condition.

Ethical and related principles are referred to as \emph{normative},
and symbolic AI systems have been developed that can represent such
principles and thereby perform ``normative reasoning''
\cite{Ciabattoni-etal:Dagstuhl-normative23}.  These have been applied
to trolley problems, including some that involve self-harm (e.g.,
throwing yourself in front of the trolley) and thereby violate the
``unintended'' aspect of DDE
\cite{Bringsjord-etal06,Govindara-etal17}.  It is claimed that fairly
sophisticated logical treatments (e.g., intensional logics,
counterfactuals, deontic modalities) are needed to represent normative
scenarios, and these might be additional to what is needed for the
primary functions of the system (hence, must be introduced explicitly,
which will add complexity).  Additionally, when normative requirements
are introduced as rules in some formal notation, there is concern that
they may have internal conflicts or otherwise lack wellformedness, and
methods have been proposed for checking this
\cite{Feng-etal:normative24}.  Additional complexities arise when
ethical judgments must be made against vague criteria (e.g., some
medical diagnoses): is the criterion uncertain (\emph{indeterminacy}
\cite{Williams:indeterminacy14}) or merely unknown
(\emph{epistemicism} \cite{Wright:vagueness95})?

Other recent work formalizes Kant's categorical imperative (humans
must be treated as ends, not as means), which requires a treatment of
causality \cite{Lindner&Bentzen18}, while others favor ``Virtue
Ethics'' and ``Artificial Phronesis'' derived from Aristotle
\cite{Vallor16,Sullins:phronesis21,Chella-etal:phronesis}.  And note
that we have mentioned only logical or rule-based representations for
ethics, whereas game theory provides another perspective.

Application of constraints derived from general normative principles
requires the guard to build a model of its world and to interpret the
principles appropriately.  It is unlikely that a non-AI perception
system can build a model that is adequate for this task, so the model
must be built by an AI perception system, or even based on that used
by the LLM itself.  We have seen previously that assurance for world
models constructed by AI-based perception systems is challenging and
currently infeasible, so only weak forms of safety assurance can be
delivered by guards based on ethics and other general rules.

And even if the AI system builds an accurate model of its world, it
may not correctly interpret its own role within that world and may
therefore be unable to apply its constraints appropriately.  For
example, the guard for an LLM that is asked to rewrite given text may
not ``know'' whether it is being used to help a non-native speaker
improve their writing, or to perform plagiarism.\footnote{The Khanmigo
``AI-powered teaching assistant'' from Kahn Academy has mechanisms to
detect this and other abuses, but it is an application built on an
LLM, not a bare LLM.  See\\
\url{https://www.youtube.com/watch?v=rnIgnS8Susg}.}  Building an
accurate and comprehensive world model is difficult in a focused
application; it is next to impossible for a general-purpose tool
lacking information on the possible contexts of its use.

In addition to ethics, AI systems should follow the laws and social
norms of their community and there is a long history of work on
formalizing and reasoning about legal systems \cite{Gardner87}.  But
there will surely be circumstances where the law conflicts with some
interpretation of ethics, or with the mission objective, so a system
constrained by several such ``overarching'' normative frameworks must
have a means of resolving conflicts.  Individually and in total, these
are challenging objectives.

We conclude that general constraints built into guards for
general-purpose AI tools cannot provide automated protection nor
assurance, largely because their correct application depends on the
guard building an accurate model of the world in which to interpret
them, or on trusting the LLM itself to provide a suitable model (which
is contrary to its generally model-free operation).  It is, however,
possible that, with suitable programming interfaces or suitable
sensitivity to prompts, the presence of general constraints could
provide useful capabilities that humans can use to add controls for
specific applications built on these general-purpose tools, albeit
with weak assurance, and this is a direction we would like to see
pursued. 

\subsection{Reputation}

Humans, endowed with good models of the world and general
understanding of local ethics and laws, sometimes make bad judgments,
or resolve conflicts among competing ethical principles in ways that
society finds unsatisfactory.  Various forms of censure and punishment
provide means to correct such errant behavior and AI systems could
also be subject to adjustment and tuning in similar ways.  An
important question then is what is the ``accounting method'' that
guides such adjustments: is it just some internal measure, or is there
some societal score-keeping that has wider significance?  In a work
commissioned by the US government during WWII, the anthropologist Ruth
Benedict proposed a distinction between ``guilt cultures'' (e.g., the
USA) and ``shame cultures'' (e.g., Japan) \cite{Benedict46}.  This
distinction is widely criticized today, but modern \emph{reputation
systems}, as employed for eBay sellers, Uber drivers, and so on can
be seen as mechanizing some aspects of shame culture\footnote{China's
Social Credit system \cite{Kobie19} extends this to the whole
society.} and could provide a framework for societal control of AI and
ML systems: the idea being that the system should be programmed to
value its reputation and to adjust its behavior to maximize this.

A necessary element in managing reputation is that it must be possible
to identify the products of a specific AI system.  For example, given
a block of text or image or video, it should be possible to tell if it
was generated by an AI tool and if so which one.  This has merit and
utility beyond management of reputation and a plausible approach has
recently been advocated \cite{Knott:detection23}.  Of course, it is
possible that some (presumably human) agents who award reputation
credits or demerits conspire to reward harmful behavior---so a whole
ecosystem, rather like the credit reporting agencies, may be necessary
to manage a reputation system.

\subsection{Checkable Outputs}

Jacovi and colleagues \cite{Jacovi-etal:trust21} discuss ``intrinsic''
and ``extrinsic'' trust in AI systems.  The latter is based on
evaluation of observed behavior and we have previously, in Section
\ref{trad}, discussed the infeasibility of deriving high levels of
assurance from observation alone.  In those discussions, the obstacle
was the vast number of tests needed for useful assurance in the
absence of justified prior confidence and the analysis required to
justify extrapolation from past to future behavior.  In the context of
general-purpose AI systems, less assurance may be acceptable and,
given some prior confidence, the number of tests might be feasible;
instead, the difficulty becomes construction and evaluation of
adequately wide-ranging tests, given that potential applications are
unknown.  Some developers of general-purpose tools are reported to
manage this using crowdsourcing.\footnote{For example, using the
Amazon Mechanical Turk: \url{https://www.mturk.com/}.}  An alternative
approach uses one AI system to test and evaluate another
\cite{Greenblatt-etal23:subversion}.  A reputation system, as sketched
above, can be seen as a way of assigning and updating extrinsic trust
in a ``live'' system.

In contrast to extrinsic trust, intrinsic trust is based on the
ability of an AI system to explain or justify its behavior.  There is
a vast amount of work on ``Explainable AI'' and we mentioned
counterfactual explanations and the fact that they can be manipulated
in Section \ref{specific}.  \emph{Constructive explanations}, an
alternative to counterfactuals, provide reasons to persuade a
``checker'' that the AI system's proposed action or output is
appropriate, or at least reasonable.  If the checker is a human, this
requires them to be sufficiently expert in the application area that
they can use the explanation to construct a chain of reasoning and
verify its accuracy.  This is unrealistic in many applications: for
example, a lay person is unlikely to be able to verify an explanation
for medical advice from an ``AI doctor.''  Recall also the problem
discussed in Section \ref{specific} of generating and interpreting
explanations when the world models of the AI system and the human
checker are misaligned.  

An alternative may be to use an automated checker as a guard: in
addition to, or instead of, a response and explanation in natural
language, the AI system could produce something like a proof, with the
proposed response as its conclusion and the explanation as its
premises, possibly together with some indication of the reasoning
purportedly employed, all in a format (e.g., SMT-LIB
\cite{Barrett-etal:SMT-lib}) that can be interpreted by an automated
checker.  The function of the checker/guard is to verify validity of
the explanation/proof and veracity of the premises (hence, soundness
of the overall response).

Since the checker will use symbolic AI (e.g., automated deduction for
validity checking and perhaps a natural language interface to
Wikipedia, or some specialized source, for fact checking), it is an
``unassured guard'' employed in a manner similar to the
``dual-process'' architecture of Section \ref{diversity} and it
provides limited assurance based on diversity and the assumption of
independent failures.

\subsection{Guardrails and Architecture Patterns}

The developers of LLMs and other general-purpose AI and ML systems are
actively working to improve their trustworthiness
\cite{Seoul:interim23,UK-AIsafety24}.  However, as we will describe in
the following section, much of the response has concerned
``existential'' risks rather than everyday dependability (or
misalignment vs.\ reliability \cite{UK-AIsafety25}).  This is
unfortunate as we think that dependability is the more urgent problem
and is becoming even more so as AI applications built on general
purpose foundations become widespread due to rapid improvements in the
performance of AI foundation models.  Ironically, we fear these
improvements may lead to \emph{less} dependable applications as those
who build applications around foundation models gain trust in them and
reduce their external checks and supervisory systems, instead building
them as ``guardrails'' on the foundation model itself.

Referring back to the architecture diagrams of Section \ref{archs},
the minimal arrangement there labeled \ref{bare} can be recast as
follows when an LLM or other foundation model is employed.\\

\includegraphics[width=0.9\textwidth]{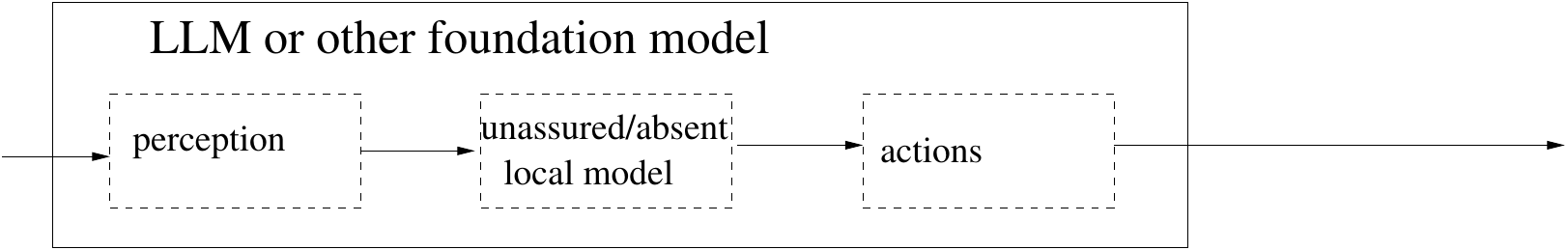} \mbox{}\\[1ex] The
idea is that the perception, local model, and action elements are all
now part of the LLM\@.  Thus, when we add ``guardrails'' similar to
the earlier Architecture \ref{actionguard}, they are likewise internal
as portrayed below.\\

\includegraphics[width=0.9\textwidth]{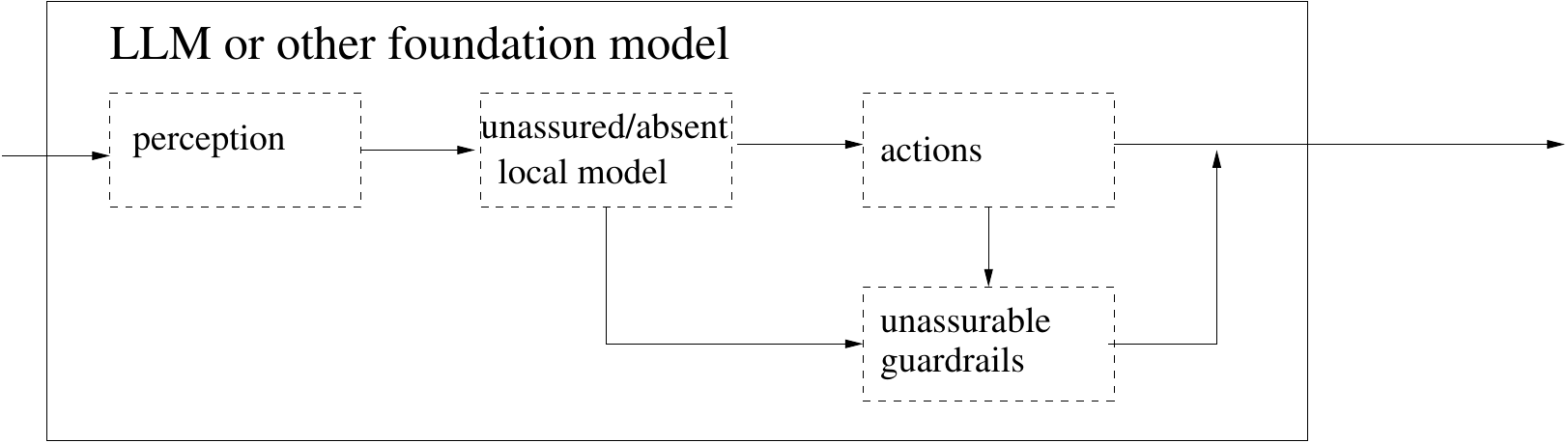}
\mbox{}\\[1ex]
Like that earlier architecture \ref{actionguard}, the guardrails are
driven from the same perception and local model as the main action
system and provide no protection for faults in those elements.  But
unlike the earlier architecture, here it is difficult to argue that
the internal guardrails are either independent or diverse.
Furthermore, there is evidence that foundation models are able to fake
their alignment
\cite{Greenblatt-etal:faking24,Meinke-etal:scheming23}.  This is not
to assert that internal guardrails cannot provide benefit, but that
the benefit cannot be assured.

We expect to see further major and perhaps surprising developments in
both capability and superficial trustworthiness of LLMs and other
general-purpose foundation models over the next few years, but we
doubt these will deliver true assurance: minor failures may become
less frequent as the foundation models improve, but there is little
reason to suppose that serious ones will decline
significantly.\footnote{\label{fattails}This is due to ``fat tails''
on the failure distribution.  Koopman provides an example: in a
variant with thin tails, we might have 100 serious faults each with an
arrival rate of 1 in 100 million demands; in a thick-tailed variant we
might have 100,000 serious faults each with an arrival rate of 1 in
100 billion demands.  Both variants expect a bad event every million
demands but after exposure or assurance equivalent to a billion
demands we will have seen all the faults of the first variant but only
1\% of the second.}  Instead, we must continue to look to external
checks as portrayed below, which is based on Architecture
\ref{dualmodels} of Section \ref{archs}.\\

\includegraphics[width=0.9\textwidth]{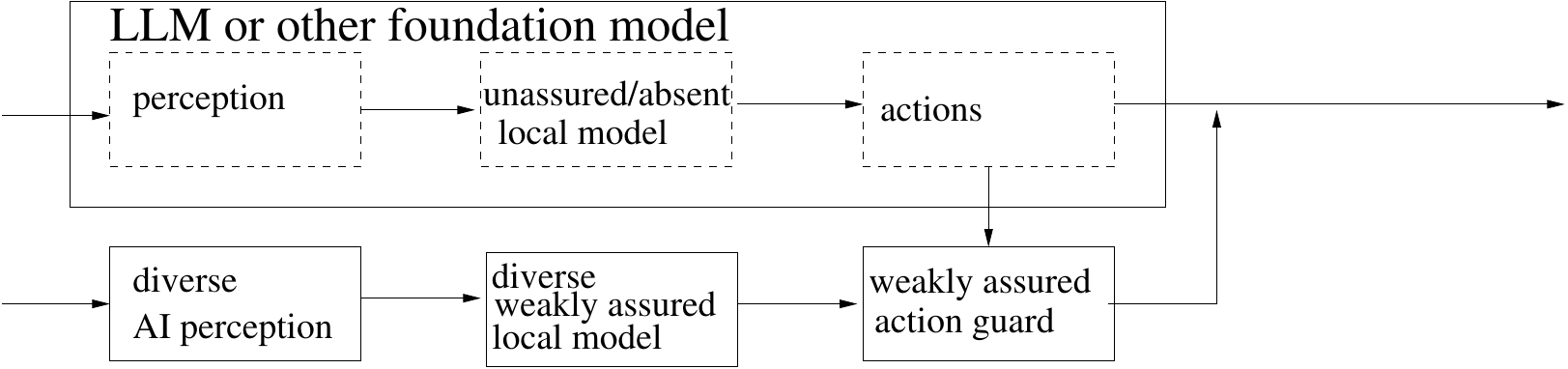}
\mbox{}\\

We believe that the methods presented here cover the feasible
approaches for some degree of dependability assurance in applications
of general-purpose systems.  In order to invoke external checks for
runtime assurance, we recommend that developers of LLMs and other
general-purpose systems provide APIs and ``hooks'' that those who
construct applications on their foundation can use to program
protections and to invoke external guards suitable for their specific
context.  These could include human review, automated review by a
diverse model, checking against normative rules, and adjustment in
response to reputation scores.  We note that some AI frameworks
already provide APIs for accessing external reasoning tools (e.g.,
\cite{Lu-etal:OctoTools25}) and it is possible that these and also
mechanisms for chain of thought reasoning could be used or adapted for
the assurance purposes advocated here.

We conclude this section by revisiting the traditional strategies for
dependability.

\begin{description}
\item[Fault avoidance.]  Although there is much merit in making the
general-purpose AI as reliable and trustworthy as possible, it is
impossible to provide assurance without knowing the application
context and potential system hazards.

\item[Fault tolerance.]  Generic guardrails can be seen as elements in
a fault avoidance strategy.  But guardrails and other checks
constructed later in the development process, when the system context
is known, can be part of a fault-tolerance strategy for systems built
around general-purpose AI\@.  For credible assurance, these components
should be external to the main AI so that claims of independence and
diversity are plausible (recall the third architecture diagram above).

\item[Failure Management and Resilience.]  The issues and challenges
discussed in this section illustrate the value of resilience
strategies.  General-purpose AI may become widely deployed in the
absence of context-specific protections, simply on the (inappropriate)
basis of interpersonal trust.  Resilience can provide some measure of
risk management and social trust.  We note that a  wide range of
potential harms must be considered and anticipated, including those of
existing and possibly regulated risks as well as general societal
harms.

\end{description}

\section{Assurance and Alignment for AGI}
\label{agi}

AGI stands for Artificial \emph{General} Intelligence and refers to
hypothetical future developments of AI and ML that can deliver human
or greater levels of performance across a wide spectrum of---and
eventually all---tasks undertaken by humans.  Beyond AGI lies the
realm of superintelligence \cite{Bostrom:superintelligence14} and
``The Singularity'' \cite{Eden-etal:Singularity13} where machines
outperform humans in all tasks, and potentially establish their own
goals without reference to human wellbeing.  Somewhat independent of
these is the possibility that machines could become conscious
\cite{Rushby&Sanchez18}.

Concerns raised by these developments, which are considered fanciful
by some but inevitable by others, are that AGI poses ``existential''
threats to humanity ranging from widespread unemployment, destruction
of social institutions, provocation of civil unrest or international
conflict, all the way to our enslavement or extinction.  Separate from
threats posed by AGI itself are concerns that near-AGI (so-called
``frontier'' models and systems) could enable bad actors to develop
catastrophic cyber, chemical, biological, radiological, and even
nuclear (CBRN) threats and attacks.\footnote{The necessary information
is assumed to be present on the Web, but dispersed and not in a form
that is useful to outsiders; but frontier LLMs may be able to
systematize, summarize, and synthesize this information into a readily
exploitable form.}

These concerns have become a topic of public and government discourse
due to very rapid recent advances in the capabilities of systems using
AI and ML\@.  In the last ten years, AI systems based on Reinforcement
Learning (RL) have beaten world champions in Chess and Go,
outperformed conventional weather-prediction models, and solved the
folding problem for proteins.  These capabilities are rightly seen by
scientists as impressive and beneficial (e.g., 2024 Nobel Prizes for
physics and chemistry) but they excited little public attention.  That
step was achieved by OpenAI's release of LLM-based ChatGPT in November
2022 \cite{OpenAI:InstructGPT22}, its adoption by Microsoft, and the
release of similar systems by Anthropic, Google and
others.\footnote{See reports of the \emph{AI Index}
\url{https://aiindex.stanford.edu}.}  LLMs generate near human-quality
text and program code in response to modest ``prompts''
\cite{openai:prompt-eng23}, and similar systems such as Stable
Diffusion and Sora generate images and videos
\cite{Xhang-etal:Sora24}.

\memotwo{Assurance for multistep actions, such as games.  How can I tell
if this move is good until I see how the whole thing plays out?}

The public is also aware of rapid increase in the capabilities of ML
for image and language recognition and in automated perception, as
seen in face recognition, language translation, voice assistants, and
self-driving cars.  Two decades ago, DARPA's ``Grand
Challenge'' for autonomous cars to drive a course in open country left
all its participants failed or crashed \cite{Burns:Autonomy18}.  Yet
today, self-driving cars are routine, if not yet fully safe.  The
general public is less aware that AI technology also performs
automated design, where ML-enabled systems rapidly generate and
evaluate designs for buildings, drugs, underwater and aerial
drones and so on, and can even propose scientific theories
\cite{Blades:AI-theorygen21} and solve open math problems
\cite{Romera-Paredes:funsearch23}.

In previous sections we have considered assurance for specific systems
and for general-purpose tools based on AI and ML where the concern is
that faulty behavior may lead to harm.  For projected developments of
current AI and ML systems and the potential emergence of AGI, however,
the concern is not just faulty behavior, nor even system failures, but
the social impact of new capabilities.  In particular, an AGI system
capable of setting its own goals might pursue objectives contrary to
human interests.  Notice that although the terminology is seldom
employed in discussion of these topics, these are nonetheless
dependability failures and can be examined from that perspective: an
AI system with potentially contrary goals is a hazard that should be
recognized and should be furnished with requirements to mitigate the
danger, together with dependability requirements analysis to show that
they do so, and a system boundary and implementation that assuredly
applies them.  However, current practice in the field frames the
assurance problem for AGI as ensuring that its behavior and goals
\emph{align} with those of human society
\cite{alignment-wiki,Gabriel:alignment20,Ji-etal:alignment20,UK-AIsafety25}---and
this alignment needs to be maintained even though AGI may fall into
the hands of bad actors, criminals, and adversaries.

\subsection{Fairly General/Good AI, AFGI}

Some consider that ``safe superintelligence is the most important
technical problem of our time'' (\url{https://ssi.inc}), but an
insidious form of disruption may arise long before AGI is available:
namely, AI systems that are good enough and cheap enough to displace
(possibly superior) human services---what we might call Artificial
Fairly General (or Fairly Good) Intelligence (AFGI, pronounced
``AFF-GEE'').  For example, well-researched journalism already finds
it hard to compete with LLM-generated press releases and propaganda
that masquerade as ``news.''  Related to this is another near-term
disruption: the possibility that ubiquitous use of LLMs and related AI
tools will reduce our collective sagacity by repeatedly circulating
misinformation, falsehoods, and mediocrity, so that we lose the
ability to access or recognize truth, novelty, or insight
\cite{Marchal-etal:AI-misuse24,Orwell:1984}.  This extends to images,
sound, and videos, where ``deepfakes'' enabled by diffusion models
invert the adage that ``seeing (and hearing) is believing.''  Beyond
the hazards of these somewhat passive, human-directed applications of
AI and ML are those of active ``agents'' driven by similar technology.
These act autonomously, performing self-determined tasks to meet
human-determined goals.  There are many possibilities for failure
here, including poorly selected tasks, misinterpreted goals, and a
misperceived environment.

In the mere six months since the first edition of this report was
issued, AFGI capabilities have developed at a remarkable rate.  One of
the more significant developments is the ``chain of thought'' (CoT)
approach to ``reasoning'' where a problem is broken in smaller chunks
that are solved sequentially \cite{Zhang-etal:CoT22}.  This was first
widely seen in OpenAI's o1 model, but is now commonplace (giving rise
to the terminology ``Large Reasoning Model,'' or LRM).  Another is use
of ``distillation'' where a powerful LLM is used to improve and fine
tune a weaker model \cite{Hintone-etal:distillation15}.  This can
generate models with performance close to the powerful LLM but small
enough to run on a single modest computer or even a mobile device.
Somewhat related is use of fewer bits to quantize weights and
parameters in the underlying neural nets: e.g., using $\{-1, 0, +1\}$
(1.58 bits) instead of 4, 8, or 16 bit integers \cite{Ma-etal:1bit}.
Again, the reduced model is much smaller but almost as good as the
original and has less need of costly high-performance GPUs.  Finally,
the open source DeepSeek R1 model that was released in January 2025
achieves performance comparable to OpenAI o1 at a fraction of the
development cost (so it is claimed).\footnote{Contrary claims assert
it is a distillation of other models.}  And in addition to LLMs, AI
corporations and cloud providers now supply components and
environments for constructing AI agents.\footnote{For example, see
\url{https://aws.amazon.com/what-is/ai-agents/}.}

If we project these advances and general research progress forward a
few years, we can expect very powerful and affordable AFGI systems to
be widely available and we propose that it is important to consider
the risks they pose in addition to those of speculated AGI systems.
(It is a controversial topic whether AFGI systems lead to AGI: there
is an empirical ``scaling law'' \cite{Kaplan-etal:scaling30} that LLMs
get better the bigger they are.  Some believe this leads inevitably to
AGI, others that (much) more is required and that LLMs are already
approaching the limits of scaling.  We subscribe to the latter opinion
because AGI surely requires the ability to reason over models of the
world and, as we have stressed several times, LLMs are ``model
free.'')

The interim report from the Seoul conference on the safety
of advanced AI identifies a broad range of
risks\cite{Seoul:interim23}.  Unfortunately, the report explicitly
chooses not to address ``Narrow AI,'' which is ``used in a vast range
of products and services\ldots{}and can pose significant risks in many
of them'' and focuses on frontier LLMs.  And even within that
coverage, the ``Seoul Commitments'' focus on ``existential risks''
initiated by malign users with access to hypothesized models with
near-AGI capabilities.\footnote{The Seoul Report states that its
restricted focus is due to ``the limited timeframe for writing this
interim report.''  The recent full report \cite{UK-AIsafety25} has
much broader coverage.}  Following the Seoul Conference, many
corporate developers of frontier AI systems signed up to the Seoul
Commitments with their focus on existential risks and have published
accounts of progress detecting and controlling
these\excite{UK-AIsafety24:commitments}.\footnote{A list of corporate
safety policies is available at \url{https://metr.org/faisc}.}  In
addition, several countries established oversight bodies (e.g., the UK
AI Safety/Security Institute, \url{aisi.gov.uk}) and these also tend
to focus on existential risks.

We, on the other hand, believe that the capabilities of current or
near-term AFGI have reached a stage where they (or systems based on
similar technology) may be deployed in ``narrow'' applications (as
discussed in Sections \ref{specific} and \ref{generic}) in preference
to custom solutions, and hence the safety risks in these applications
should be considered part of advanced AI safety.  This means we must
consider component failures initiated by faults within the system
(e.g., ``hallucinations'') and ``normal accidents'' due to system
failures, as well as those initiated in the environment (e.g., by
rogue users), and we should consider general dependability and
creeping disability as well as existential risks.

We examined dependability in applications and in general-purpose
models such as LLMs in the previous two sections and we do not think
availability of AFGI frontier models changes our assessments, other
than the scale and urgency of the risk---although we are concerned
that the apparent quality of these models will encourage construction
of ``guardrails'' on or within the frontier model itself, rather than
externally, where diversity and independence can provide rather
more---but still modest---assurance.  And we are also concerned that
the fluency of AFGI systems (for images, speech, and video, not just
language) will embed them deeply into social contexts so that the
system boundary is unknown and low-level system failures may become
endemic and serious ones become possible.

Between failures and existential risks there is a category of societal
harms and disruptions that AFGI may precipitate.  First, there are
unfortunate consequences precipitated by the training of frontier
models.  LLMs and diffusion models are trained by scraping text,
images, and all manner of other material from the Web.  This likely
violates copyright in many instances\footnote{New York Times, 27
December 2023.} and jeopardizes the livings of writers, artists, and
musicians and their associated ecosystems as LLMs plagiarize their
work.  Furthermore, as LLMs become widely used, their own output
becomes a large part of the web corpus that trains the next generation
and so on, potentially causing the Web to become full of LLM-generated
pablum.  There already is evidence for this: replying to a prompt
asking it to tinker with malware, an LLM called Grok, developed by
xAI, refused ``as it goes against OpenAI's use case policy.''  OpenAI
has nothing to do with xAI, so Grok presumably generated this response
by scraping web text generated by OpenAI's LLMs.  This phenomenon is
termed \emph{Model Collapse} \cite{Shumailov-etal:model-collapse24}
and is predicted to ``dumb down'' the overall information content of
the Web.  Meanwhile, search and recommender systems for music, video,
and books may lead us to serendipitous and enjoyable discoveries, but
they can also be manipulated to channel our attention along
predetermined paths.

Another consequence of training is massive use of electric power and
water resources by the servers employed, potentially harming the
environment.  Government and corporate initiatives are proposing to
spend huge sums on development and use of additional server farms (the
``Stargate'' project targets half a trillion dollars,\footnote{see
\url{https://openai.com/index/announcing-the-stargate-project/}.}
or approximately the GDP of Norway or Malaysia).  In addition to
environmental harm, the costs involved may limit development, and
ultimately control, of frontier technology to just a few countries and
corporations, so that an oligarchy of magnates could subvert the
structure of established societies and their governments.

Moving from training to deployment, a generally anticipated disruption
is widespread unemployment or other labor market disruption when AFGI
capabilities little better than those available today displace office
jobs ranging from clerks to middle-management, together with skilled
jobs such as software coding and routine design, and professions such
as lawyer and doctor.  These AFGI capabilities might be superior to
human performance, or they might be inferior but cheaper and ``good
enough''; either way, they could displace human jobs.  Beyond harm to
individuals, unfettered use of AFGI, particularly in conjunction with
polarization perpetrated by social media, could undermine our
institutions, or our trust in these.  All these disruptions could be
chronic rather than acute: that is, they could develop over a period
(e.g., creeping unemployment) rather than abruptly and, as with
climate change, this may make recognition difficult, and effective
response hard to mobilize.

In addition to cases where harm may be unintended, there is the
threat that AFGI could multiply the capabilities of adversaries and
hackers (e.g., by automating the search for vulnerabilities and the
coding of viruses to exploit them, or the generation of phishing
emails, provocative social media posts, and hurtful ``deepfake''
images and videos).  In the hands of adversaries, AFGI-enabled
disinformation campaigns and social media manipulations could sow
societal discord and conflict.  Until recently, interventions of this
kind required the capabilities and resources of a nation state, but
are now coming within reach of small groups or lone actors.

Then there are truly dystopian applications of AFGI such as ``Lethal
Autonomous Weapons Systems'' (LAWS, aka.\ ``killer robots'').  Most
countries have policies that require some form of human control or
approval over these (e.g., USA \cite{CRS-LAWS24}) but it is unknown
how these limitations will hold up in combat.  And there are AI
systems that contribute to lethality without being weapons.  For
example, AI-powered intelligence systems identify vastly more
potential targets more quickly than humans can (e.g., 100 a day vs.\
50 a year \cite{AI-targeting23}).  By doctrine, humans are required to
select and approve actual targets from those identified, but the sheer
numbers make meaningful oversight difficult.

A very short ``consensus'' paper with 24 distinguished authors
outlines several other potential risks of AI and proposes various
technical and governance measures \cite{Bengio-etal:risks23}.  In that
regard, the UK held the first AI Safety Summit at Bletchley Park in
2023 and established an AI Safety Institute (AISI) that performs
``rigorous AI research to enable advanced AI governance''
(\url{https://aisi.gov.uk}).  Likewise, on 30 October 2023 the United
States Government issued Executive Order 14110 on the ``Safe, Secure,
and Trustworthy Development and Use of Artificial Intelligence.''  The
Executive Order directed the National Institute of Standards and
Technology (NIST) to ``develop guidelines and best practices to
promote consensus industry standards that help ensure the development
and deployment of safe, secure, and trustworthy AI systems'' (see
\url{https://www.nist.gov/aisi}), but this order was revoked by the
successor government on 20 January 2025.  The UK subsequently renamed
its AISI as the AI \emph{Security} Institute, and the UK and the USA
declined to endorse a declaration on ``inclusive and sustainable'' AI
that was signed by 58 other countries at the February 2025 AI Action
Summit in Paris.

The European Union has a recent (May 2024) law on AI that takes a
risk-based approach to AI applications, categorizing them in four
levels: minimal, limited, high, and unacceptable.  In addition, for
the first time, it proposes to regulate the underlying technology of
foundation models and General-Purpose AI (GPAI).  Foundation models
must comply with transparency obligations, and ``high-impact models
with systemic risk'' will have to conduct model evaluations, assess
and mitigate systemic risks, conduct adversarial testing, report to
the European Commission on serious incidents, ensure cybersecurity and
report on their energy efficiency.  GPAIs with systemic risk may rely
on codes of practice\footnote{See
\url{https://artificialintelligenceact.eu/introduction-to-codes-of-practice/}.}
to comply with the new regulation.

Most of these steps seem well-intentioned\footnote{A contrary point of
view is that the focus on existential risks is a form of regulatory
capture (by distraction), allowing AI corporations unregulated freedom
to perpetrate other risks \cite{Khlaaf:time23}.}  but apart from the
EU's GPAI proposals, we consider them unlikely to provide much
benefit.  Every novel systemic societal hazard posed by recent
computer systems was unanticipated and unrecognized until the harm was
done (they can be seen as system failures),\footnote{Merton
\cite{Merton:consequences36} was the first to explicitly identify
unanticipated consequences, which later became more often termed
``unintended'' consequences; Zwart \cite{Zwart15} criticizes this
transition.}  and subsequent regulation has proved ineffective or
counterproductive.  For example, the advent of advertising as the main
revenue source for Web services has led to universal surveillance of
personal online activity, to the extent that even our cars invade our
privacy.\footnote{See
\url{https://foundation.mozilla.org/en/privacynotincluded/categories/cars/}.}
This was not anticipated and regulatory responses such as ``cookie
warnings'' are a source of annoyance rather than protection, and
indicate technological illiteracy on the part of regulatory agencies
and lawmakers.  Similarly, the poisonous impact of ``algorithmic''
traffic generation on social media was not anticipated and still has
no effective response.

\subsection{Resilience as a Key Response to AFGI/AGI}

The emergence of AFGI and potentially of AGI requires an eco-system
for impact monitoring including citizen groups, government agencies
such as AISI, international NGOs, and risk owners and their
regulators.  We suggest that, in addition to monitoring AI technology,
these entities should monitor its impact on society, so that
unanticipated risks and incipient system failures can be detected
early, and addressed.  Monitored risks should include chronic and
creeping developments such as accumulation
\cite{Kasirzadeh:existential24} and gradual disempowerment
\cite{Kulveit-etal:disempowerment25}.  In addition to technological
adjustments and assurance, risks should be mitigated through societal
adaptation for, as we have already noted several times, safety and
assurance are system properties and with deployment of AFGI the system
can extend widely into its societal context.  Thus, in parallel with
mechanisms to protect against the risks of AFGI, we also need measures
to make our systems and society more \emph{resilient} to those risks.
So, for example, if the hazard of unemployment is to be mitigated,
there should be research and action on how humans and AI can
productively work \emph{together}, with widespread training on best
practices.

We also suggest that hazard analysis should be performed according to
best practices (which will need further development to deal with the
unique hazards of AI) and that the upper levels of risk should be more
finely graduated than simply ``existential.''  True existential risk
refers to the extinction of humanity or civilization as we know it,
comparable to nuclear war.  There are many levels of serious and
catastrophic risk below this and they require graduated levels of
monitoring and assurance, with special care to ensure they do not
accumulate.  (Although it is focused on climate change, a report of
the IPCC provides useful general descriptions and definitions for
extreme risk and risk management together with resilience strategies
such as coping, incremental adjustment, transformation, and adaptation
\cite{IPCC:risks12}.)  Catastrophic system failures often accumulate
from a combination or succession of smaller faults and abnormalities
within a complex system (recall Perrow's ``normal accidents'') and
safety is best achieved by making systems more resilient through being
able to break and work around failure pathways (e.g., by avoiding what
Perrow calls ``tight coupling'' and ``interactive complexity''
\cite{Perrow84}) rather than striving to eliminate all component and
subsystem failures.

Furthermore, it does not seem widely appreciated in the AI community
that assurance does not scale linearly with risk: the cost and effort
and the techniques employed for assurance of critical avionics, for
example, are totally different than those for less critical systems,
to the extent that the assurance for a Level A avionics system
typically costs more than development of the system.  Yet we see
assessment for CBRN hazards in LLMs performed by ``red teams.''  This
may suitable of initial exploration, but assurance against the
asserted magnitude of this risk requires vastly more effort (and/or
societal adaptation, such as surveillance, controlled access to
precursors, development of protective measures, antidotes, hardening
of targets, and so on).

In our opinion, a vital means for anticipation and control of
potential disruptions due to AI is self-imposed or mandatory scrutiny
\emph{within} the organizations developing the technology,\footnote{As
precedent, the FAA has Designated Engineering Representatives (DERs)
within the aviation industry: the argument being that only those
actually working on the products know enough of the details to
identify and anticipate hazards.  This is obviously vulnerable to
regulatory capture, as demonstrated by the Boeing MCAS scandal, but is
now reinforced and has otherwise worked well.}  combined with public
disclosure and debate leading to establishment of regulatory
goals.
We believe there is a strong role for modern methods such as assurance
cases and goal-based regulation within this framework, rather than the
``codes of practice'' envisioned in the EU regulations, especially for
systems that perform specific functions.  The reason that other
industries have moved from standards and codes of practice to explicit
goals and assurance cases is hard-won experience that assurance has to
be based on identification and elimination of the hazards of the
specific system and environment under consideration.  For failures
precipitated by AFGI systems, such as advanced LLMs, we believe that
mechanistic protections such as guards and internal monitors may prove
inadequate and that new methods will need to be developed based on
improved understanding of topics such as system failure, emergent
behavior \cite{Kopetz-etal:emergence16}, the cognitive basis of
language and shared intentionality (see below), and the sociology of
judgment and cooperation \cite{Irving&Askell19:AI-socsci}.

\subsection{True AGI}

So far, our discussion has considered AFGI systems that are not far
beyond those already current, and has not touched on ``true'' AGI\@.
We suspect that popular opinion misinterprets advanced AFGI as AGI and
thereby underestimates how far we are from true AGI and its potential
dangers.  We have already stated our opinion that true AGI requires
more than the facility with language exhibited by AFGI: it requires
the ability to \emph{think}
\cite{Fedorenko-etal24:lang-comms-not-thought}, which we interpret as
construction and manipulation of general models of the world, and this
is not yet in sight.  And there is something apart from General
Intelligence that makes humans the masters of planet Earth.  A single
human is a puny thing and no threat to plants or animals, but
collectively we wreak our will.  And it is not just force that is
multiplied by our teamwork: we also increase our collective
intelligence.  Individually, we do not develop pottery, writing,
mathematics---those are products of collective culture.  Computers
communicate and can be programmed to act collectively, but this is not
the same as cultivating alliances, sharing knowledge, ``selling'' an
idea, and ultimately forming a team with shared goals and coordinated
plans: that is \emph{shared intentionality}
\cite{Tomasello:origins-communication10,Tomasello&Carpenter07,Bickerton:Adam09,Rushby:SIFT23}
and it is unique to humans.\footnote{Social animals like ants, bees,
wolves, dolphins exhibit collective behavior, but this is ``programmed
in'': a wolf cannot cause its pack to adopt a \emph{new} idea.}  Thus,
until AGI achieves shared intentionality, it is no great threat to
humanity by itself (although in the hands of bad actors it could be):
we can always act collectively to outwit it.  Of course, this does
depend on our willingness to act collectively and history is replete
with instances of societies that thought they could exploit a
dangerous entity, yet retain control (e.g., Germany in 1933).

As widely quoted (and variously attributed) ``it is
difficult to make predictions, especially about the future,'' and we
consider the technology and the timing of true AGI to be
unknowable.\footnote{But see the \emph{One Hundred Year Study on
Artificial Intelligence} \url{https://ai100.stanford.edu/}.}  Hence,
it is impossible to identify potential hazards and suitable
mitigations and methods of assurance.  However, it does seem sensible
to consider possible scenarios, even speculative ones (so science
fiction may be an appropriate forum to perform this work).  For
example, as already noted, one dystopian possibility is that AGI may
misinterpret its goals or establish its own at variance to our desires
and best interests, or that we may simply give it poorly chosen goals.
Hadfield-Menell and colleagues \cite{Russell-etal:CIRL16} provide
examples of poorly specified goals (e.g., King Midas' wish that
everything he touches turn to gold) or misinterpreted ones (e.g., a
robot given the goal of cleaning up dirt repeatedly dumps and cleans
the same dirt).  They identify the generic problem as one of
\emph{value alignment} and propose ``Cooperative Inverse Reinforcement
Learning'' (CIRL) as a promising technical framework, later
generalized to ``assistance games'' \cite{Russell:assistance-games21}.
The idea is that the AGI's goal is continuously to learn what it is we
want it to do by observing our behavior (but what if its ``master''
has bad intent?).  A more wide-ranging and speculative proposal is for
``Guaranteed Safe AI'' \cite{Dalrymple-etal24:GS-AI}, which we outline
and criticize in \cite[pages 7, 8]{Bloomfield&Rushby:FAISC24}.

One thing we can note is that, as in all the previous sections of this
report, an AGI system will surely operate by building some model of
its world and will then formulate goals and actions based on that
model.  In previous sections we have seen that it is generally
possible to construct some guarding function to mitigate the hazards
of harmful goals or faulty actions, but this guard will also depend on
a world model.  Thus the fundamental problem in dependable AI and AGI
systems is assurance for perception and model construction.
Dependability requirements will be stated in terms of human models of
the world (e.g., three dimensions of space and one of time) and in
order to enforce them, a guard's model will presumably need to use the
same framework.  Yet AGI capability may derive from alternative
frameworks (i.e., selection of latent variables)\footnote{There are
arguments going all the way back to Kant that our perceived model of
the world cannot correspond to ``reality'' \cite{Hoffman:reality19}.}
and alignment of human and machine models built on different
frameworks seems a substantial challenge.

Beyond AGI is the concern that machines may become conscious
\cite{Rushby&Sanchez18}.  This is a difficult topic because there is
no agreed definition for consciousness, nor ways of detecting or
measuring it (consider debate over whether animals are conscious).
And AI consciousness may exist but be utterly unlike our own (for an
analogy, consider the possibility of a consciousness distributed among
the nine ``brains'' of an octopus
\cite{Carls-Diamante22,Mather:octopus-mind19}).  Nonetheless, there is
general agreement on two aspects of consciousness: \emph{intentional}
consciousness\footnote{The term is a translation from German and
should not be considered to focus on ``intentions.''} is the ability
to direct attention and to think \emph{about} something and to know
that you are doing so, while \emph{phenomenal} consciousness is ``what
it's like'' to have subjective experiences such as the smell of a rose
or the feeling of pain.  There are two subtopics here: one is how
plausible or likely it is that AI could achieve either kind of
consciousness, and the second is what the consequences might be.

We should note that humans generally attribute behaviors to
intentional consciousness that actually originate in the subconscious
\cite{Libet85} and, contrary to our intuitions, the conscious mind is,
for the most part, less an initiator of actions and more a reporter
and interpreter of actions and decisions initiated in the subconscious
\cite{Gazzaniga:who12}.  Hence, when we see complex behaviors in
animals, we attribute it to intentional consciousness because we
falsely believe that is how it is with us.  Speech in humans
does require consciousness, so we are apt to anthropomorphize
systems with language skill and impute consciousness to them even
though they contain no mechanisms that could plausibly generate it
(cf.\ ELIZA of 1966 \cite{Weizenbaum:Eliza66} and also Footnote
\ref{lemoine}).

Having said that, there is no agreement how human consciousness is
achieved, nor what purpose it serves: Kuhn describes a ``landscape''
with more than 200 theories of consciousness
\cite{RLKuhn24:consc-landscape}.  However, there have been
experimental attempts to construct machine consciousness
\cite{Reggia13,Gamez08}.  Mostly, these are based on the idea that
consciousness derives from explicit models of self and of others, and
is related to communication and language
\cite{Holland03,Pipitone-etal19}: these ideas correspond most
closely with Higher Order Thought (HOT) models of human consciousness
\cite{Brown-etal19,Gennaro04,Rushby:SIFT23}.  None of these
experiments have delivered any sign of consciousness, although they
have sharpened some of the questions, and the possibility of machine
consciousness remains open.

Were it to be achieved, intentional AI consciousness seems unlikely to
add capabilities beyond those of AGI: intentional consciousness seems
to be essential to human reasoning, but AI produces facsimiles of
reasoning by other means, just as airplanes fly without growing
feathers or flapping their wings.  However, AI intentional
consciousness might require us to attribute some of the system's
behavior to conscious decisions, thereby raising philosophically
difficult questions regarding moral responsibility, personal identity,
and free will.

Phenomenal consciousness would also raise philosophically difficult
questions, but in this case they would concern ethics---not what AI
might do to us, but how we should treat AI\@.  An AI system that truly
has subjective experience, that feels pain and pleasure, raises
profound questions on the foundations of ethics and on just treatment
of other sentient beings
\cite{Butlin&Lappas:AI-consc26,Birch:edge-sentience24}.\footnote{Several
researchers have signed open letters urging caution
\url{https://amcs-community.org/open-letters/} and proposing five
principles
\url{https://conscium.com/open-letter-guiding-research-into-machine-consciousness/}
regarding research and development of potentially conscious AI
entities.}  It also asks how we could tell whether the AI's subjective
experience is real or faked (cf.\ ``philosophical zombies''
\cite{sep-zombies}).

\section{Summary and Conclusion}

We have described and discussed methods to ensure and assure critical
properties of AI systems from the traditional ``dependability''
perspective.  This requires that assurance for critical properties
rests only on elements of the system (which may be ``guards'' rather
than operational functions) for which we have near-complete
understanding of what they do, how they do it, why they do it, and the
environment in which they do it, and comparably complete understanding
of their implementation and its correctness.  This understanding is
needed because testing and operational experience on their own are
inadequate for strong assurance: they need to be buttressed by prior
confidence in the quality of the system and arguments that the future
product and environment will behave like the past.  All of these
topics should be addressed in an assurance case that assembles claims,
evidence, arguments, and theories in a manner that provides sufficient
confidence in its top (i.e., overall) claim to justify deployment.

The dependability perspective asserts that systems based on AI and
Machine Learning (ML) cannot satisfy the requirements for assurance
based on detailed understanding because their inner working is opaque;
the contrary ``trustworthy'' perspective believes they can, in some
cases.  There is a continuum or spectrum between these viewpoints and
all are worthy of investigation; we focus here on those at the
dependability end and hope others will survey other points along the
spectrum.

In that regard, we note recent papers that propose to apply assurance
or safety cases to the trustworthy perspective (e.g.,
\cite{Clymer24:AI-safetycases,Dalrymple-etal24:GS-AI}).  We welcome
these developments but stipulate that when we speak of an assurance
case we set a high bar \cite{Bloomfield&Rushby:Assurance2}.  In
particular, the case must be \emph{indefeasible}, meaning there is no
credible new information that would change its assessment
\cite{Rushby:Shonan16} and this should be probed skeptically by
exploring potential defeaters \cite{Bloomfield-etal:defeaters24}.  A
(deliberate) consequence is that the case must have ``no gaps,'' which
generally requires the argument to be deductive
\cite{Bloomfield&Rushby:confidence22,Bloomfield&Rushby24:CBJ};
exceptions must be noted as \emph{residual doubts} and shown to pose
negligible risk.\footnote{Koopman \cite[Section
7.2]{Koopman:howsafe22} asserts that assurance cases for complex
systems such as self-driving cars cannot be deductive because they
operate in an open world that cannot be fully predicted.  We respond
that they can still be deductive but some of their elements may be
incompletely characterized.  The difference is between forcing
consideration of weaknesses and doing something about them, such as
monitoring at runtime or accepting them as residual risks (i.e.,
``known unknowns'') versus allowing the argument to have unconsidered
gaps (i.e., ``unknown unknowns'').}  Evidence must be assessed
skeptically (we advocate use of confirmation measures
\cite{Good:weight83,Rushby:AAA13}) and shown to support not merely a
claim of \emph{something measured} (e.g., ``we did penetration testing
to such and such standard'') but one of \emph{something useful} (e.g.,
``therefore faults of type xx are absent'').  In well-developed
technical areas, the assurance argument should largely be assembled
from standard \emph{theories} that provide well-attested (and ideally
``pre-certified'') subcases for common fragments of the overall case
(e.g., static analysis for absence of certain coding faults).

The upper levels of the assurance case do not concern the design or
implementation of the system, but a description of its environment and
assumptions, development of its requirements, and derivation of its
hazards, all bound together by dependability requirements validation
to demonstrate that the requirements mitigate the hazards.  Next comes
the specification that describes the design of the system, coupled
with intent verification to show that these satisfy the requirements.
Only then do we get to the implementation of the system and its
correctness verification where, we stress, for critical systems we
need near certainty that this satisfies the specification.  We suggest
that formal verification of properties of neural networks often
corresponds more closely to static analysis that full correctness
verification.

Due to these challenging stipulations, we do not consider that
assurance for trustworthy AI behavior is feasible in most current
applications, although we welcome continued research in these
directions and look forward to positive developments
\cite{Bloomfield&Rushby:FAISC24}.  We also acknowledge that many AI
applications do not require the same degree of assurance as
safety-critical cyber-physical systems (CPS), but it is challenging to
determine which parts of a full assurance case may then be relaxed,
and how, and by how much (probabilistic estimation may be useful
\cite[Section3]{Bloomfield&Rushby:confidence22}).
Work by Dong and colleagues \cite{Dong-etal:MLassurance24} meets many
of our stipulations but much of the evidence for their assurance case
comes from statistical modeling of the reliability of its ML
components.  This is reasonable from a trustworthiness perspective but
less so from the dependability viewpoint, where we would want to see
runtime verification and dynamic assurance monitoring.

The dependability viewpoint does not trust AI and ML elements because
they are derived by experimental optimization (``training'') and their
complete behavior is unknown.  Consequently, assurance is achieved
using runtime verification with guarded architectures in which overall
behavior is checked for required properties by traditionally developed
and assured guards.  An obstacle to this approach is that the guards
must accurately perceive (i.e., build a model of) the current state of
the system's environment, and this itself may require AI and ML (e.g.,
for a self-driving car to perceive other road users).

We examined this dilemma in some detail for CPS extended with AI and
ML\@.  In some cases (e.g., ``geofencing'') adequate perception can be
achieved by traditional means, and in some others traditional
mechanisms can support crude but moderately safe guarding functions
(e.g., emergency braking in self-driving cars).  We then considered
whether AI perception could be beneficial in some circumstances and we
examined the topics of diversity and defense in depth and concluded
that moderately assured architectures could be constructed around
these ideas, primarily to reduce demands on traditionally engineered
backup guards.

We then considered other kinds of systems that perform specific
functions and make use of AI and ML\@.  Knowing the function performed
by a system, it is generally feasible to identify its hazards and
critical properties, and we then found that similar approaches and
architectures to those examined for CPS can be feasible, although
there may be novel challenges in assurance for both perception and
guards.

We then considered the idea of system failures (or ``normal
accidents'') which are failures that are not (primarily) due to
component faults but to unanticipated interactions among correct
components and subsystems.  Because AI systems have facility with
language and other modes of human dialog, they can become more deeply
embedded into their human social context than intended or recognized.
This means that the system boundary is not well defined and makes
system failures more likely.

Next, we looked at foundation models such as LLMs and other
general-purpose systems on which more specific systems can be
constructed (previously these required custom AI components).  We
acknowledged the possibility of system failures but focused on
techniques for guarding general-purpose AI and ML mechanisms.  Here,
the additional problem is that the ultimate application and its
hazards are unknown and so the general-purpose system has to use very
general guards.  We considered guards based on normative principles,
such as ethics and other overarching frameworks, and we also examined
explanations and reputation systems.  Again, we found that effective
guards may be feasible, but the main difficulty is assurance that the
world view constructed by the guard aligns with human perception.

Finally, we considered AGI and other futuristic prospects such as the
singularity and machine consciousness.  Here, the concern is less
about faults and failures of AI systems and more about system failures
and other unanticipated and disruptive consequences of their correct
and intended behavior.  It is infeasible to guard these systems so we
briefly examined ways to ensure that their goals \emph{align} with
ours (which, again, is largely a question of how the world is
perceived).  Although we concede that near-term development of AGI is
possible, we suggest that a more imminent concern is AFGI (Artificial
Fairly General Intelligence), which can be projected from current
advances in generative AI such as ``frontier'' LLMs, and we identified
some proposed government regulations and other safeguards.

Many of these are concerned with ``existential'' risks that we
consider inadequately defined: we would like to see far more detailed
hazard analysis and more granular (as opposed to ``all at once'')
scenarios for their emergence.  We note that current methods of
investigation and assurance (basically ``red-teaming'') are woefully
incommensurate with the claimed seriousness of the risk.\footnote{A
recent evaluation reported that the best performing system (OpenAI o1)
failed 26\% of tests while the worst (DeepSeek R1) failed 100\%:
\url{https://blogs.cisco.com/security/evaluating-security-risk-in-deepseek-and-other-frontier-reasoning-models}.\\
The acceptable failure rates in any critical system (i.e., one that
poses serious risks) are many orders of magnitude more demanding than
this.}  Proposed mitigations are comparably weak, such as not
disclosing the weights \cite{Nevo-etal:weights24} in the neural nets
of LLMs considered to pose ``significantly higher risk''
\footnote{Anthropic has a four-level scale for AI Safety Levels
(ASLs): ASL-1: smaller models, ASL-2: present large models, ASL-3:
significantly higher risk, ASL-4: speculative.}
\cite{Anthropic:RSP24}; distillation can largely copy the capabilities
of an LLM without access to its weights.  Hence our call for more
precise hazard analysis followed by rational determination of the
claims that need to be assured and the confidence required in their
assurance.  These will require new or revised hazard and failure
analysis methods that address the unique aspects of AI and ML (and
AFGI in particular): for example, and as mentioned earlier, augmented
guidewords for those hazard analysis methods that use them
\cite{Crawley&Tyler:HAZOP15,McDermid96}.

These methods should address hazards and mitigations over the full
development lifecycle, from those focused on growth of foundation
models, through the construction of applications built around them, to
customers and general society that are impacted by their use.  We
recommend that mitigations and assurance should focus on the overall
system, not just the AI mechanism, and this will extend into human
society, where resilience to grave risks should be developed alongside
measures for their prevention.

Overall, we suggest that for CPS and other systems for specific
functions, architectures similar to those labeled \ref{combo},
\ref{allin} and \ref{modds} in Sections \ref{archs} and \ref{odds} are
(in ascending order) the most attractive.  These use a traditionally
engineered and assured guard for ``last second'' protection
(\ref{combo}), plus weakly assured but diverse perception for defense
in depth (\ref{allin}), and assured detection of micro-ODDs with
specialized guards for each one (\ref{modds}).  Similar architectures
can be adapted for general-purpose systems, but with less assurance
(as the hazards are less well-defined and perception more difficult).
For AFGI, and even more so for AGI, we suggest that protection and
alignment must go beyond restrictions on their mechanisms and build on
an understanding of system failures and emergence, the cognitive basis
of language and intelligence, and the sociology of group cooperation.

\subsection{Research Agenda}

There are four research topics that we consider urgent.
\begin{itemize}

\item Methods of hazard analysis are needed that can identify
potential sources of failure and harm in complex AI systems, together
with design and assurance methods to avoid such harms.  Hazard
analysis needs to consider the new harms and failure modes of AI
systems, their propensity to embed themselves deeply into their social
context and thereby widen the system boundary, potentially leading to
self-organizing criticality, and system failure (``normal
accidents'').

In addition to hazards, research should examine the credible severity
of risks due to AI and ML and should consider chronic and creeping
harms as well as catastrophic ones.  Research should also distinguish
those risks whose mitigation primarily requires technological measures
and regulation from those that also require societal response and
adaptation.

\item Development of layered, recursively structured architectures and
associated guards to provide runtime verification within a
socio-technical approach to design (i.e., one that integrates
technology with human and community aspects) that avoids both
component and system failures.

It is unlikely that guards alone can provide full protection for an AI
system in a complex environment, so overall architectures should favor
defense in depth based on diversity and other rational principles,
which begin with careful appraisal of what the system is to do, and
the environment in which it will operate.

In addition to architectures and to design and assurance methods that
aim to avoid failure, research should explore credible methods for
dynamic assurance and post-failure resilience of systems and their
social context.

\item Assurance for models of the world/environment/context built by
perception systems using AI and ML,

Much current work focuses on development of guards, but these are
dependent on accurate perception of the world and we consider that
assured perception is a neglected topic.  This applies not only to
explicit local models (e.g., location of other vehicles) as
constructed by CPS, but implicit ones (e.g., current social context
and role within it) underlying the behavior of LLMs.

Assured perception systems need not provide fully detailed local
models, but must be accurate to the level of resolution or
discrimination needed for modest but effective guards (e.g., detection
of micro-ODDs and prediction of urgent actions).  Research from
the trustworthiness perspective could usefully be applied to this
problem.  Related to assured perception is the topic of principled
fusion of diverse local models, possibly having different levels of
resolution (e.g., one fully detailed, and another less detailed but
assuredly accurate).  A promising approach uses ``predictive
processing'' and a ``dual process'' cognitive structure similar to the
human brain.

\item Research on the computational mechanisms underlying emergent
behavior and the cognitive basis of language, intelligence and shared
intentionality,

We cannot develop methods for assurance and alignment of AFGI and AGI
without better understanding of the computational processes underlying
cognition including the development and use of domain models.  Human
(and to some extent animal) cognition are the only models that we
have, so techniques such as use of metaphors to map one domain (e.g.,
social hierarchy) onto another more automated one (e.g., height)
\cite{Lakoff&Johnson08-old} may repay study.  Beyond cognition, we
need to understand the processes underlying teamwork (i.e., shared
intentionality) and the sociology of cooperative behavior.

\end{itemize}

Finally, we recommend development of concrete instances of assured
perception systems and of recursively structured guarded
architectures, and their theoretical and practical evaluation.

\paragraph{Acknowledgments.}

We are grateful for constructive and insightful comments received from
readers of previous versions of this report, particularly Phil Koopman
(CMU), Brian Randell and Cliff Jones (Newcastle), and Wilfried Steiner
(TTTech).

This material is partially based on work supported by the United
States Air Force and DARPA under contract FA8750-23-C-0519.  Any
opinions, findings and conclusions or recommendations expressed in
this material are those of the author(s) and do not necessarily
reflect the views of the United States Air Force and DARPA.

\newcommand{\doi}[1]{\href{https://doi.org/#1}{\tt DOI:#1}}
\section*{References\markboth{References}{References}}
\addcontentsline{toc}{section}{References}
\bibliographystyle{modplain}

\end{document}
% End of foo.tex -----------------------------------